\documentclass[lettersize,journal]{IEEEtran}
\usepackage{amsmath,amsfonts}
\usepackage{algorithmicx}
\usepackage{algorithm}
\usepackage{algpseudocode}
\usepackage{array}
\usepackage[caption=false,font=footnotesize,labelfont=rm,textfont=rm]{subfig}
\usepackage{textcomp}
\usepackage{stfloats}
\usepackage{url}
\usepackage{verbatim}
\usepackage{graphicx}
\usepackage{cite}
\usepackage{epsfig}
\usepackage{enumitem}
\usepackage{booktabs}
\usepackage{diagbox}
\usepackage{siunitx}
\usepackage{makecell}
\usepackage{bbding}
\usepackage{multirow}
\usepackage{colortbl}
\usepackage{bm}
\usepackage{soul}
\usepackage{ifpdf}
\ifpdf
  
\else
\fi

\newcommand{\eg}{\textit{e}.\textit{g}.}
\newcommand{\ie}{\textit{i}.\textit{e}.}

\usepackage{color}
\definecolor{cyan}{cmyk}{1,0,0,0}
\definecolor{darkgreen}{rgb}{0,0.5,0}
\definecolor{orange}{rgb}{1,0.5,0}
\definecolor{magenta}{cmyk}{0,1,0,0}
\definecolor{darkyellow}{cmyk}{0,0,0.75,0}
\definecolor{gray}{rgb}{0.8,0.8,0.8}
\definecolor{lightgray}{gray}{0.9}
\definecolor{lightblue}{rgb}{0.5,0.5,0.8}
\usepackage{hyperref}
\newcommand{\HalfCheck}{\Checkmark\kern-1.2ex\raisebox{1ex}{\rotatebox[origin=c]{125}{\textbf{--}}}}
\graphicspath{{figures/},{pairs/}}

\begin{document}
	
	\title{YOND: Practical Blind Raw Image Denoising\\Free from Camera-Specific Data Dependency}
	
	\author{Hansen~Feng, 
		Lizhi~Wang,~\IEEEmembership{Member,~IEEE,}
		Yiqi~Huang,
		Tong~Li,
		Lin~Zhu,~\IEEEmembership{Member,~IEEE,}\\
		and~Hua~Huang,~\IEEEmembership{Senior Member,~IEEE}
		\IEEEcompsocitemizethanks{
			\IEEEcompsocthanksitem Hansen Feng, Yiqi Huang, Tong Li and Lin Zhu are with the School of Computer Science and Technology, Beijing Institute of Technology, Beijing, 100081, China. Email: $\{$fenghansen, huangyiqi, litong, linzhu$\}$@bit.edu.cn\protect
			\IEEEcompsocthanksitem Lizhi Wang and Hua Huang are with the School of Artificial Intelligence, Beijing Normal University, Beijing, 100875, China. Email: $\{$wanglizhi, huahuang@bnu.edu.cn$\}$\protect
		}
	}
	
	\markboth{Journal of \LaTeX\ Class Files,~Vol.~14, No.~8, August~2021}%
	{Shell \MakeLowercase{\textit{et al.}}: A Sample Article Using IEEEtran.cls for IEEE Journals}
	
	\IEEEpubid{0000--0000/00\$00.00~\copyright~2021 IEEE}
	
	\maketitle
	
	\begin{abstract}
		The rapid advancement of photography has created a growing demand for a practical blind raw image denoising method. Recently, learning-based methods have become mainstream due to their excellent performance. However, most existing learning-based methods suffer from camera-specific data dependency, resulting in performance drops when applied to data from unknown cameras.
		To address this challenge, we introduce a novel blind raw image denoising method named YOND, which represents You Only Need a Denoiser. Trained solely on synthetic data, YOND can generalize robustly to noisy raw images captured by diverse unknown cameras.
		Specifically, we propose three key modules to guarantee the practicality of YOND: coarse-to-fine noise estimation (CNE), expectation-matched variance-stabilizing transform (EM-VST), and SNR-guided denoiser (SNR-Net).
		Firstly, we propose CNE to identify the camera noise characteristic, refining the estimated noise parameters based on the coarse denoised image.
		Secondly, we propose EM-VST to eliminate camera-specific data dependency, correcting the bias expectation of VST according to the noisy image.
		Finally, we propose SNR-Net to offer controllable raw image denoising, supporting adaptive adjustments and manual fine-tuning.
		Extensive experiments on unknown cameras, along with flexible solutions for challenging cases, demonstrate the superior practicality of our method. The source code will be publicly available at the \href{https://fenghansen.github.io/publication/YOND}{project homepage}.
		
	\end{abstract}
	
	
	\section{Introduction}\label{sec:introduction}
	\IEEEPARstart{W}{ith} the widespread adoption of mobile devices and the opening of raw data interfaces~\cite{API}, an increasing number of individual users desire to experience advanced photography techniques. Inevitably, diverse camera noise always reduces the effectiveness of these techniques in practice, highlighting the need for a practical blind raw image denoising method.
	Recently, learning-based methods~\cite{CVPR18/SID,CVPR19/Unprocess,ECCV20/Yuzhi,CVPR20/ELD,TPAMI21/ELD,ICCV21/SFRN,IJCV22/zhangyi,TPAMI23/PMN,ICCV23/LED,CVPR23/LLD,CVPR22/IDR,TPAMI24/VDN} have become mainstream due to their excellent performance.
	Unfortunately, learning-based methods trained on data from specific cameras often struggle to generalize to other unknown cameras~\cite{ECCV20/Yuzhi, ICCV23/LED}. We attribute this problem to camera-specific data dependency, which presents a serious challenge in developing a practical blind raw image denoising method.
	\begin{figure}[t]
		\centering
		\includegraphics[width=1.0\linewidth]{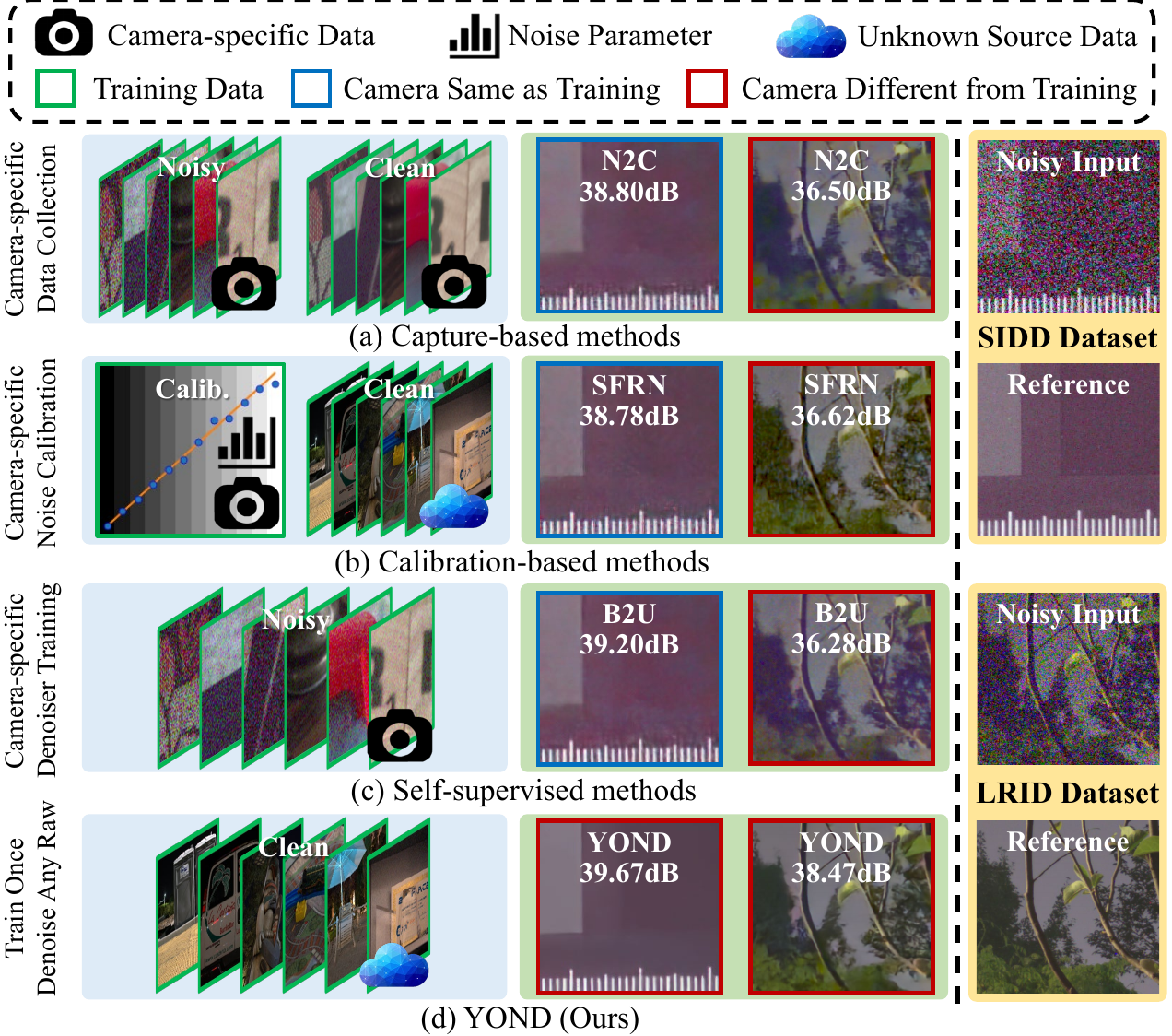}
		\caption{Comparison of data dependency and denoising performance across various raw image denoising methods. Due to camera-specific data dependency, capture-based~\cite{CVPR18/SIDD}, calibration-based~\cite{ICCV21/SFRN}, and self-supervised~\cite{CVPR22/B2U} methods struggle to effectively generalize to noisy data captured by unknown cameras. In contrast, YOND exhibits superior performance on data from unknown cameras after once training, which demonstrates the practicality of our method. \textbf{(Best viewed with zoom-in)}}
		\label{fig:teaser}
	\end{figure}
	
	According to the camera-specific data dependency, we can divide the learning-based raw image denoising methods into three categories. 
	Firstly, capture-based methods~\cite{CVPR18/SID,TPAMI23/PMN,ICCV23/LED,CVPR23/LLD} depend on paired real data to supervise the network training, which involves camera-specific data collection. However, data defects (\eg, residual noise, spatial and brightness misalignment) are commonly prevalent in public raw image denoising datasets~\cite{CVPR17/DND,CVPR18/SIDD,TPAMI23/PMN,CVPR24/MIDD}, causing methods trained on such data to overfit to specific cameras.
	\IEEEpubidadjcol
	Secondly, calibration-based methods~\cite{P-G,ECCV20/Yuzhi,CVPR20/ELD,TPAMI21/ELD,ICCV21/SFRN,ICME21/RethinkNM} depend on calibration data (\eg, flat-field frames and dark frames) to synthesize the training data, which involves camera-specific noise calibration. 
	However, unknown camera characteristics and environmental variations introduce uncertainties, which causes a mismatch between the calibrated results and real-world noise distributions~\cite{TPAMI23/PMN,ICCV23/LED}. 
	Lastly, self-supervised methods~\cite{ICML18/N2N,CVPR21/FBI,CVPR21/NBR2NBR,CVPR22/B2U,CVPR22/IDR,CVPR23/MIT} depend on independent identically distributed (i.i.d.) noisy data to approximate the supervised training, which involves camera-specific denoiser training. However, additional constraints must be introduced to compensate for the idealized i.i.d. assumption, which leads to expensive training costs and suboptimal denoising performance.
	In conclusion, as shown in Figure~\ref{fig:teaser}, existing methods are suffering from various practicality problems caused by camera-specific data dependency. Therefore, it is important and necessary to break camera-specific data dependency for blind raw image denoising.
	
	In this work, we introduce a novel practical blind raw image denoising method free from camera-specific data dependency. 
	Our method begins by estimating the noise distribution from a single image to identify the camera noise characteristic. Next, we apply a variance-stabilizing transform (VST) to eliminate camera-specific data dependency by normalizing the noise into additive white gaussian noise (AWGN). Finally, we employ an adaptive AWGN denoiser guided by the estimated signal-to-noise ratio (SNR) to deliver both precision and flexibility. \textbf{Trained solely on synthetic data, our method can effectively generalize to noisy data captured by diverse unknown cameras}. We name our method YOND, as you need nothing else under our method, \textbf{Y}ou \textbf{O}nly \textbf{N}eed a \textbf{D}enoiser.
	
	Specifically, we propose three key modules to guarantee the practicality of YOND: the coarse-to-fine noise estimation (CNE), the expectation-matched variance-stabilizing transform (EM-VST), and the SNR-guided denoiser (SNR-Net).
	
	Firstly, we notice that existing noise estimation methods~\cite{P-G,2013ICIP/Liu,CVPR21/FBI} have limited precision, resulting in unstable after-VST noise level and subsequent denoising performance. To address this problem, CNE begins by estimating coarse noise parameters directly from noisy images, then refines these parameters by collaboratively leveraging the noisy images and their corresponding coarse denoised images. Our CNE provides high precision in noise parameter estimation, which ensures robust denoising performance.
	
	Secondly, we notice that VST~\cite{ANSCOMBE,GAT,TIP11/VST} often exhibits expectation bias, especially in low-light conditions. The existing unbiased inverse transform~\cite{TIP11/O-VST,PGD-VST12,TIP13/VST} is operated based on denoised images, which amplifies the error caused by inaccurate denoising. To address this problem, EM-VST first calculates the bias function of VST, then matches the expectation bias before the inverse transform according to the noisy image. Our EM-VST exhibits low error in VST expectation bias correction, which ensures the exact color of denoised images.
	
	Lastly, we notice that a highly controllable denoiser \cite{CVPR19/DNI,CVPR19/AdaFM,ECCV20/CResMD,PR24/CFMNet} is necessary to meet diverse challenges in practice, while most methods ignore this demand. To address this problem, SNR-Net efficiently utilizes explicit SNR for adaptive denoising and allows flexibly adjusting guidance value (\ie, SNR) to satisfy custom visual preferences in practice. Furthermore, we introduce a novel iterative strategy to extend SNR-Net into a simplified diffusion model~\cite{NIPS20/DDPM,ICLR21/DDIM,DMID} to recover image details. Our SNR-Net offers controllable raw AWGN denoising, enabling adaptive adjustment to deliver clear denoised images.
	
	We conduct comprehensive experiments on public datasets and real-world scenarios. YOND demonstrates superior performance free from depending on camera-specific data.
	Moreover, we discuss various solutions for handling challenging cases in practice. YOND offers high flexibility for manual adjustments without additional training.
	Extensive experiments on unknown cameras and flexible solutions for challenging cases demonstrate the superior practicality of our method.
	
	Our main contributions are summarized as follows:
	
	\begin{enumerate} [leftmargin=0.82cm]
		\item[1.] We introduce YOND, a practical method for blind raw image denoising free from camera-specific data dependency, which shows generalization to data from diverse unknown cameras once training on synthetic data.
		
		\item[2.] We propose a coarse-to-fine noise estimation, providing high precision in noise parameter estimation, which ensures robust denoising performance.
		
		\item[3.] We propose an expectation-matched VST, exhibiting low error in VST expectation bias correction, which ensures the exact color of denoised images.
		
		\item[4.] We propose an SNR-guided denoiser, offering controllable raw AWGN denoising, which delivers clear denoised images under adaptive adjustment.
		
	\end{enumerate}
	
	\section{Related Works}
	\subsection{Blind Raw Image Denoising} 
	Blind raw image denoising refers to the process of denoising an image without any camera-specific prior information. As a practical challenge, blind raw image denoising has been extensively studied in traditional methods, which is generally composed of noise estimation, VST, and AWGN denoiser. 
	Some supervised blind denoising methods~\cite{CVPR19/CBDNet,Pseudo-ISP,TPAMI24/VDN} employ neural networks to estimate noise variance maps to guide the denoising process. However, these methods possess ``blind" capabilities only within the limited noise level range of the same camera, and they struggle with blind denoising for images captured by unknown cameras.
	In recent years, various self-supervised blind denoising methods have been proposed, leveraging deep image prior~\cite{2018deep,liu2023devil,jo2021rethinking,CVPR23/MIT}, deep noise prior~\cite{ICML18/N2N,CVPR21/NBR2NBR,zero,ICCV23/DCD}, and blind-spot network~(BSN)~\cite{CVPR19/N2V,CVPR22/B2U,CVPR22/AP-BSN,LGBPN,CVPR24/AT-BSN,TCSVT24/CompBSN,laine2019high}. However, these self-supervised methods face a common challenge: the difficulty of effectively generalizing to data with distributions different from the camera-specific training data~\cite{CVPR23/MIT}. While some self-supervised methods~\cite{ICLR20/nnVST} facilitate single-image training, individually training a network for each image is impractical.
	The noise characteristics of raw data are favorable, allowing for a significant reduction in data dependency when combined with traditional methods.
	FBI~\cite{CVPR21/FBI} incorporates a PGE-Net for noise estimation, seamlessly integrates VST, and employs the blind-spot network FBI-Net for denoising. However, FBI-Net relies on a self-supervised BSN to compensate for inaccurate noise estimation, which consequently limits its generalizability to camera-specific training data.
	
	YOND efficiently integrates traditional blind denoising methods with neural networks. We achieve superior denoising performance while breaking the camera-specific data dependency, providing a practical solution for blind raw image denoising.
	
	\begin{figure*}[h]
		\centering
		\includegraphics[width=1.0\linewidth]{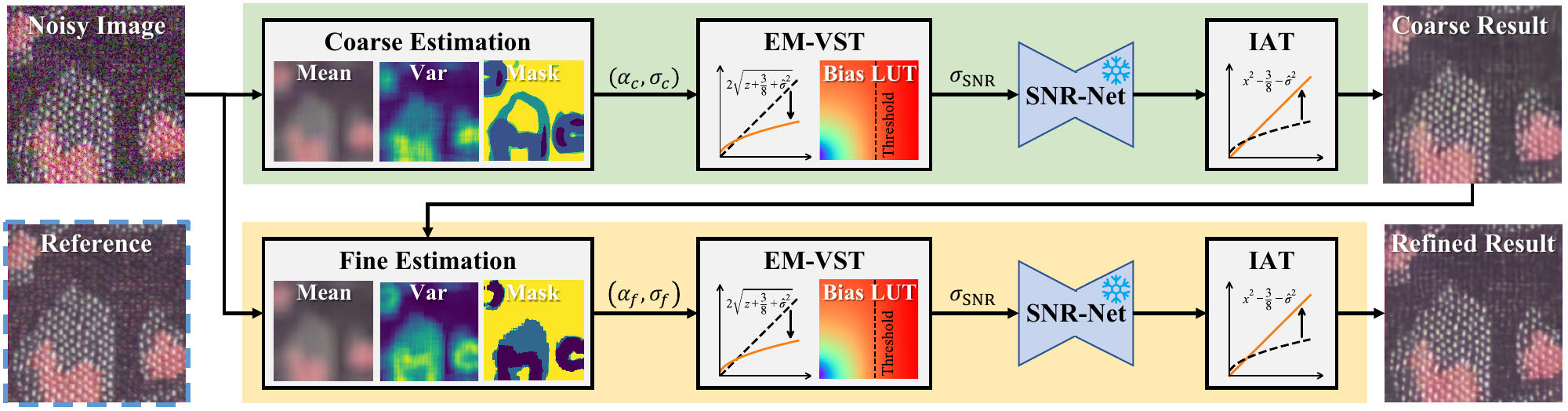}
		\caption{The blind raw image denoising pipeline of YOND. SNR-Net is our pre-trained controllable raw AWGN denoiser, which denoises the transformed noisy image under the guidance of the noise level $\sigma_{\text{SNR}}$. We visualize raw images and mean maps as RGB images via a simple ISP. We visualize masks and variance maps after packing, averaging, and normalizing.}
		\label{fig:pipeline}
		\vspace{-.2in}
	\end{figure*}
	
	\subsection{Noise Estimation}
	Raw images, without unknown nonlinear transformations caused by ISP, typically manifest stable and well-defined noise characteristics. In camera electronics, sensor noise is generally modeled as Poisson-Gaussian noise~\cite{P-G}. The sensor noise parameters can be accurately calibrated in specific environments, supporting downstream noise modeling~\cite{TPAMI94/CCD, P-G, TIP17/PMM, CVPR20/ELD, TPAMI21/ELD, TPAMI23/PMN} and image denoising ~\cite{NLM, BM3D, TIP18/FFDNet, CVPR19/CBDNet, CVPR21/FBI}. However, practical scenarios may pose challenges in capturing original raw images (\eg, RGBW sensors). Moreover, objective factors such as capture settings and physics environments somewhat impact the practicality of calibration~\cite{TPAMI23/PMN}. To address these practical challenges, noise estimation is widely applied.
	
	Traditional methods estimate noise parameters by identifying flat regions \cite{meer1990fast, immerkaer1996fast}. Introducing PCA, \cite{TIP12/PCA} effectively separates noise components from patches. \cite{ICIP13/Liu} iteratively select patches with weak textures for estimation. \cite{P-G, TIP21/DWTNE} optimize the selection of flat regions by introducing wavelet transforms, concentrating most noise signals in the high-frequency regions for improved estimation. In recent years, learning-based methods~\cite{TIP18/FFDNet,CVPR19/CBDNet,CVPR21/FBI,CVPR22/FG-ELD,ICCV23/DCD} are proposed in noise parameter estimation. Some supervised methods~\cite{TIP18/FFDNet,CVPR19/CBDNet} utilize neural networks to predict noise variance maps. However, supervised methods are essentially equivalent to additional deep supervised denoising, lacking generalizability. \cite{CVPR22/FG-ELD} can estimate fine-grained noise model parameters through contrastive learning. Nevertheless, relies on abundant camera-specific noise images, lacking practicality.
	
	We propose a novel coarse-to-fine noise estimation method combined with learning-based denoisers. Precise noise estimation constitutes the key to breaking data dependency in YOND.
	
	\section{You Only Need a Denoiser}
	\subsection{Pipeline}
	Our method builds upon traditional VST-based denoising methods~\cite{PGD-VST12,TIP13/VST,BM3D} with single-image noise estimation~\cite{P-G,ICIP13/Liu}, which are known for their strong generalizability but limited by underdeveloped modules. To improve denoising performance via learning-based methods while remaining free from camera-specific data dependency, we propose YOND (\textbf{Y}ou \textbf{O}nly \textbf{N}eed a \textbf{D}enoiser), a novel blind raw image denoising method that integrates the strengths of both traditional and learning-based methods.
	
	YOND consists of three key modules: coarse-to-fine noise estimation (CNE), expectation-matched VST (EM-VST), and SNR-guided denoiser (SNR-Net). 
	Firstly, We notice that existing noise estimation methods~\cite{P-G,2013ICIP/Liu,CVPR21/FBI} typically fails on texture scenes, thus we propose CNE to improve noise estimation precision in a coarse-to-fine manner. 
	Secondly, We notice that VST~\cite{ANSCOMBE,GAT,TIP11/VST} often exhibits expectation bias in low-light conditions, thus we propose EM-VST to reduce the expectation bias based on the statistical analysis. 
	Lastly, We notice that controllable denoiser~\cite{CVPR19/DNI,CVPR19/AdaFM,ECCV20/CResMD,PR24/CFMNet} is necessary to meet diverse challenges in practice, thus we propose SNR-Net to support flexible solutions under adaptive adjustment.
	
	The pipeline of YOND is shown in Figure~\ref{fig:pipeline}. 
	In the first stage, coarse noise parameters $(\alpha_c, \sigma_c)$ are estimated from the noisy image $y$, which is then transformed by EM-VST to yield an transformed noisy image $f(y)$ with (approximately) standard AWGN. SNR-Net then takes transformed noisy image $f(y)$ and its noise level $\sigma_{\text{SNR}}$ to produce a transformed denoised image $f(\hat{x}_c)$. This result is subsequently transformed back to the original intensity space using the inverse Anscombe transform (IAT)~\cite{TIP11/VST}, yielding a coarse denoised image $\hat{x}_c$.
	In the second stage, both of the coarse denoised image $\hat{x}_c$ and noisy image $y$ are used to refine the noise estimation, yielding fine noise parameters $(\alpha_f, \sigma_f)$, which marks the end of CNE. The process is then repeated, yielding a fine denoised image $\hat{x}_f$. It is worth noting that the coarse denoised image $\hat{x}_c$ is only used for noise estimation and does not directly contribute to denoising.
	
	With our developed pipeline, YOND shows superior generalization to data from diverse unknown cameras once training on synthetic data.
	
	\subsection{Coarse-to-fine Noise Estimation}\label{sec:CNE}
	\subsubsection{\textbf{Principle}}
	The imaging sensor noise is typically modeled as Poisson-Gaussian noise~\cite{P-G}. For noisy images $y$ and corresponding clean counterparts $x$, the model is expressed as
	
	\begin{equation}
		y \sim \alpha \mathcal{P}\left(\frac{x}{\alpha}\right) + \mathcal{N}(0, \sigma^2)
	\end{equation}
	where $\alpha$ represents the system gain of camera, $\sigma^2$ is the variance of read noise, $\mathcal{P}(\cdot)$ and $\mathcal{N}(\cdot)$ represents the Poisson and Gaussian distribution respectively.
	
	Based on the statistical properties of the Poisson and Gaussian distributions, the expectation and variance are given as
	\begin{equation}
		\begin{cases}
			&E(y) = \alpha \cdot \dfrac{x}{\alpha} + 0 = x \\ &Var(y) = \alpha^2 \cdot \dfrac{x}{\alpha} + \sigma^2 = \alpha x + \sigma^2.
		\end{cases}
	\end{equation}
	
	Noise calibration typically involves collecting extensive camera-specific calibration data~\cite{TPAMI94/CCD,EMVA1288,TPAMI21/ELD}, computing the mean and variance under varying brightness levels, and fitting the parameters $\alpha$ and $\sigma$ using a least-squares method.
	
	Following noise calibration principles, traditional noise estimation methods usually segment flat regions using techniques such as wavelet transforms and PCA~\cite{meer1990fast,immerkaer1996fast,P-G,TIP12/PCA,ICIP13/Liu,TIP21/DWTNE}. Subsequently, noise parameters are fitted based on the spatial mean and variance maps of these flat regions. However, these methods face two main challenges: (1) the threshold selection for natural image segmentation is challenging, often resulting in excessive outliers during fitting, and (2) natural images rarely contain perfectly flat regions, causing spatial variances to exceed true values.
	
	
	Building on calibration methods, we identify an approach to address these challenges. According to the linearity property of variance, for two independent random variables \(X\) and \(Y\), the variance of their sum equals the sum of their variances, \ie, \(Var(X+Y) = Var(X) + Var(Y)\). If the clean image \(x\) were known, noise estimation errors could be further reduced. To approximate this condition, CNE regards the coarse denoised image $\hat{x}_c$ as a substitute for clean image \(x\) and subtracts its variance from that of the noisy image during the fine estimation stage. This coarse-to-fine approach significantly improves noise estimation precision, ensuring robust denoising performance.

	\begin{figure*}[t]
		\centering
		\includegraphics[width=\linewidth]{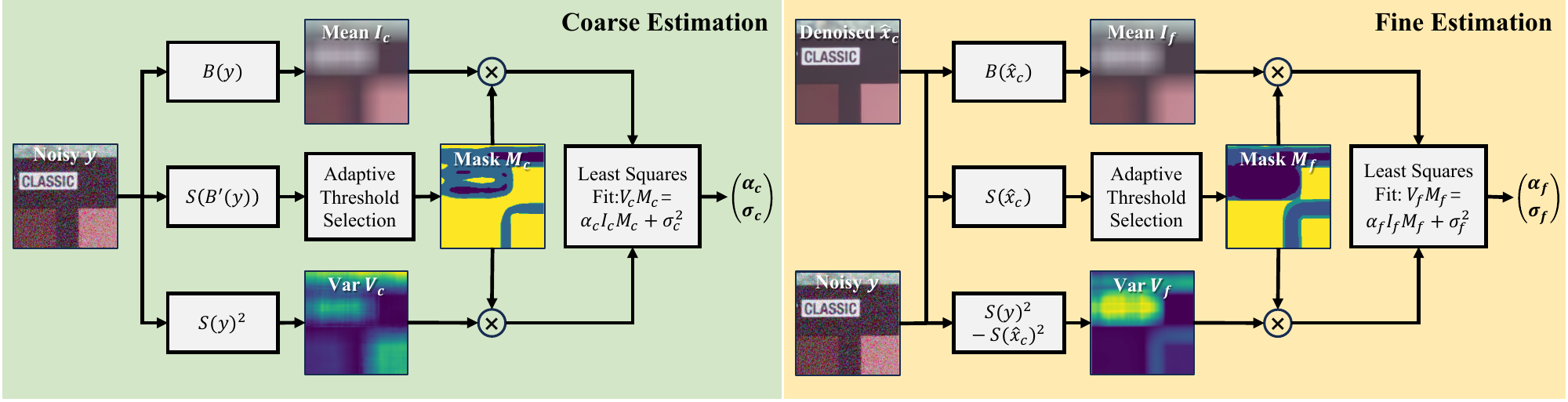}
		\caption{Pipeline of CNE with key steps visualization. The operation \(B\) denotes the 29$\times$29 mean filtering $B_{29}$. \(B'\) refers to 19$\times$19 mean filtering $B_{19}$. \(S\) represents 29$\times$29 spatial standard deviation computation $S_{29}$. For raw images and mean maps $I$, we applied a simple ISP to visualize them as RGB images. For masks $M$ and variance maps $V$, we visualize them after packing, averaging the four color channels, and normalizing.}
		
		\label{fig:CNE_pipe}
	\end{figure*}
	
	\subsubsection{\textbf{Procedure}}
	

	To simplify the description of CNE, we will define two fundamental operations. \(B_p\) denotes mean filtering with a kernel size of \(p\), used for spatial expectation calculation and noise suppression, which is defined as
	\begin{equation}
		B_p(D) = D \ast \frac{1}{p^2} \mathbf{1}_{p \times p},
	\end{equation}
	where \(D\) is the input image, \( \mathbf{1}_{p \times p} \) is a \(p \times p\) matrix with all elements equal to 1, and \( \ast \) denotes the convolution operation.
	
	\(S_p\) denotes spatial standard deviation computation with a kernel size of \(p\), used for spatial variance computation and mask threshold calculation, which is defined as
	\begin{equation}
		S_p(D) = \sqrt{B_p(D^2) - B_p(D)^2}.
	\end{equation}
	
	We visualize the pipeline of CNE in Figure~\ref{fig:CNE_pipe}.
	In the coarse estimation stage, the flat region mask \(M_c=\text{ATS}\left(S_{p}(B_{p{'}}(y))\right)\) is calculated by our adaptive threshold selection (ATS) algorithm. Flat regions are then identified on the mean map \(I_c=B_{p}(y)\) and variance map \(V_c=S_{p}(y)^2\). The coarse noise parameters \((\alpha_c, \sigma_c)\) are subsequently estimated by the least squares method. 
	In the fine estimation stage, the mean map \(I_f\), variance map \(V_f\), and flat region mask \(M_f\) are jointly calculated by the noisy image \(y\) and the coarse denoised image \(\hat{x}_c\). Specifically, \(M_f\) is generated as \(\text{ATS}\left(S_{p}(\hat{x}_c)\right)\), and flat regions are selected from \(I_f=B_{p}(\hat{x}_c)\) and \(V_f=S_{p}(y)^2-S_{p}(\hat{x}_c)^2\). The fine noise parameters \((\alpha_f, \sigma_f)\) are finally estimated by the least squares method. 
	
	To ensure computational efficiency, mean filtering is implemented with integral images~\cite{boxfilter}, allowing the runtime to remain independent of the kernel size \(p\). In practice, we set \(p=29\) and \(p'=19\) to effectively reduce noise-induced estimation variance.  
	
	It is worth noting that a robust and adaptive threshold selection algorithm is crucial for noise estimation. Our ATS leverages an intuitive prior that flat regions in an image usually exhibit low spatial variance. The prior transforms a flat region segmentation task to an adaptive single-threshold selection task.	
	To complete the algorithm, we introduce two key constraints: (1) the flat region must contain a sufficient number of pixels to meet statistical sampling requirements, and (2) the flat region should cover a diverse range of signal values to prevent overfitting during parameter estimation. 
	Based on these constraints, we define a simple optimization objective:
	
	\begin{equation}\label{eq:ATS}
		\theta = \underset{\theta}{\arg\min} \left(\frac{\theta}{q(\theta) n(\theta)}\right),
	\end{equation}
	where \(\theta\) represents the spatial variance threshold for segmenting flat regions. \(q(\theta)\) denotes the quantile corresponding to \(\theta\), indicating the percentage of pixels in the selected flat region. \(n(\theta)\) reflects the number of non-empty bins in the histogram, representing the diversity of signal values within the flat region at the given threshold.
	
	In practice, ATS evaluates 20 candidate solutions (\ie, 20 quantiles of the data distribution) to determine the optimal threshold.
	The algorithm is simple yet robust, significantly improving the precision of noise parameter estimation while maintaining computational efficiency.
	
	\begin{figure}[t]
		\subfloat[High-signal region]{
			\includegraphics[width=0.48\linewidth,trim=10 10 10 10,clip]{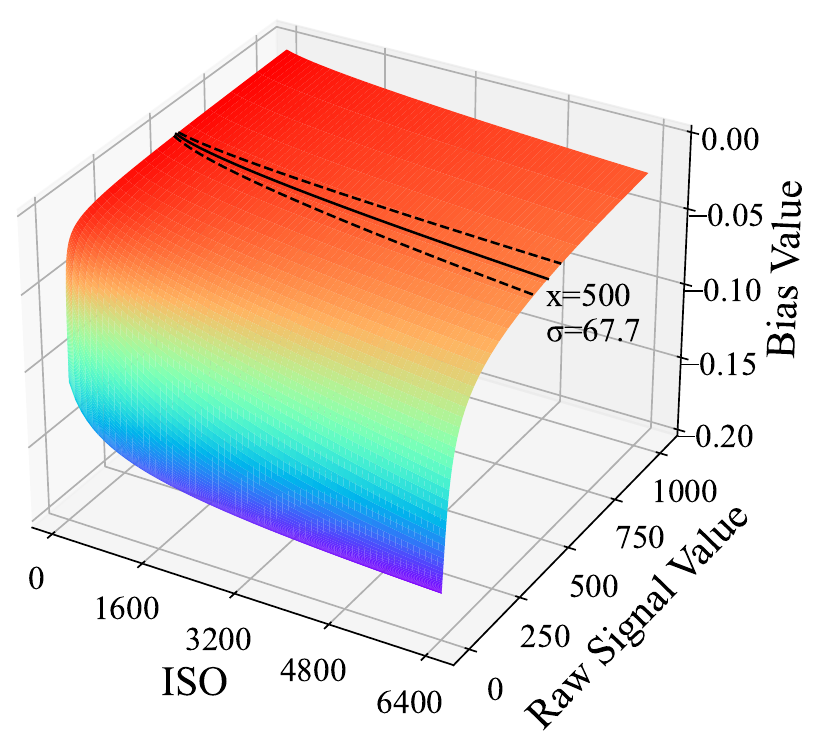}
		}
		\subfloat[Low-signal region]{
			\includegraphics[width=0.48\linewidth,trim=10 10 10 10,clip]{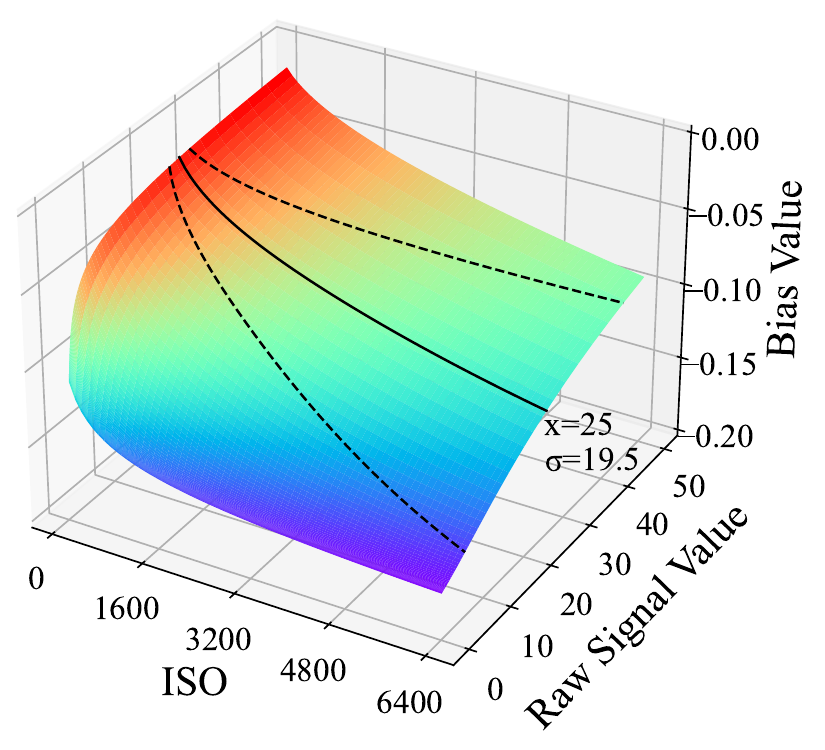}
		}
		\caption{The VST bias function $e_{\hat{\sigma}}(x)$ as a function of signal values at different ISO settings for a smartphone with an IMX686 sensor. The curves in (a) show the variation of noise level $\sigma$ with ISO settings when $x$=500, with $\sigma$=67.7 at ISO-6400. The curves in (b) show the variation of noise level $\sigma$ with ISO settings when $x$=25, with $\sigma$=19.5 at ISO-6400.\label{fig:bias_3d}}
		
		\label{fig:bias3d}
	\end{figure}
	
	\subsection{Expectation-Matched VST} 
	\subsubsection{\textbf{Principle}}\label{sec:gat}
	VST aims to stabilize the variance of a random variable to a constant through numerical mapping. In raw image denoising, VST transforms the noise distribution of each pixel in the noisy image \(y\) to (approximately) a standard normal distribution \(\mathcal{N}(0,1)\). The most classic VST is the generalized Anscombe transform (GAT) \cite{GAT}, which transforms each pixel \(y > -\frac{3}{8}\alpha-\frac{\sigma^2}{\alpha}+\mu\) to
	
	\begin{equation}\label{eq:GAT_ori}
		f(y)= \frac{2}{\alpha} \sqrt{\alpha y+\frac{3}{8} \alpha^2+\sigma^2-\alpha \mu},
	\end{equation}
	
	Real-world camera noise is generally considered to have a zero mean after black level correction, thus we simplify Eq.~\eqref{eq:GAT_ori} by letting \(z = \frac{y - \mu}{\alpha}\), \(\hat{\sigma} = \frac{\sigma}{\alpha}\), and \(\mu = 0\)~\cite{TIP13/VST}:
	
	\begin{equation}\label{eq:GAT}
		f_{\hat{\sigma}}(z)= \begin{cases}2 \sqrt{z+\frac{3}{8}+\hat{\sigma}^2}, & z>-\frac{3}{8}-\hat{\sigma}^2 \\ 0, & z \leq-\frac{3}{8}-\hat{\sigma}^2\end{cases}.
	\end{equation}
	
	As Eq.~\eqref{eq:GAT} is a nonlinear transformation, applying GAT to the noisy image results in biased expectations:
	
	\begin{equation}\label{eq:neq}
		f^{-1}_{\hat{\sigma}}\left(E\left(f_{\hat{\sigma}}(z) \mid x\right)\right) \neq x,
	\end{equation}
	where \(f_{\hat{\sigma}}^{-1}(x) = x^2 - \frac{3}{8} - \hat{\sigma}^2\) represents the IAT. The expectation \(E\left(f_{\hat{\sigma}}(z) \mid x\right)\) can be expanded as
	
	\begin{equation}
		\begin{aligned}\label{eq:intE}
			E & \left(f_{\hat{\sigma}}(z) \mid x\right)=\int_{-\infty}^{+\infty} f_{\hat{\sigma}}(z) p(z \mid x, \hat{\sigma}) d z \\
			& =\int_{-\infty}^{+\infty} 2 \sqrt{z+\frac{3}{8}+\hat{\sigma}^2} \sum_{k=0}^{+\infty}\left(\frac{x^k e^{-x}}{k ! \sqrt{2 \pi \hat{\sigma}^2}} e^{-\frac{(z-k)^2}{2 \hat{\sigma}^2}}\right) d z.
		\end{aligned}
	\end{equation}
	
	Eq.~\eqref{eq:neq} indicates that even with perfect denoising of the transformed noisy image, the VST inherently introduces a bias between the inversed image $f^{-1}_{\hat{\sigma}}\left(E\left(f_{\hat{\sigma}}(z) \mid x\right)\right)$ and the clean image $x$. The bias introduced by VST is particularly pronounced in low-signal regions, severely impacting low-light denoising performance. Therefore, it is necessary to correct the bias \(\epsilon_x\) caused by VST according to the bias function \(e_{\hat{\sigma}}(x)\):
	\begin{equation}\label{eq:bias}
		\epsilon_x=e_{\hat{\sigma}}(x)=E \left(f_{\hat{\sigma}}(z) \mid x\right) - f_{\hat{\sigma}}(x).
	\end{equation}
	
	However, computing Eq.~\eqref{eq:bias} requires access to the clean image $x$, which is impractical in denoising tasks. The closed-form unbiased inverse Anscombe transform (UIAT)\cite{TIP13/VST} attempts to address this by treating the denoised result as the clean image to correct the bias during the inverse transformation. Unfortunately, the denoised results inevitably contain residual errors, especially in low-signal regions. Consequently, the subsequent UIAT amplifies these errors, causing color bias in low-light conditions. Moreover, the image priors in low-signal regions may be distorted due to bias, while some generative restoration models are sensitive to such distortions.
	
	\begin{figure}[t]
		\centering
		\includegraphics[width=\linewidth]{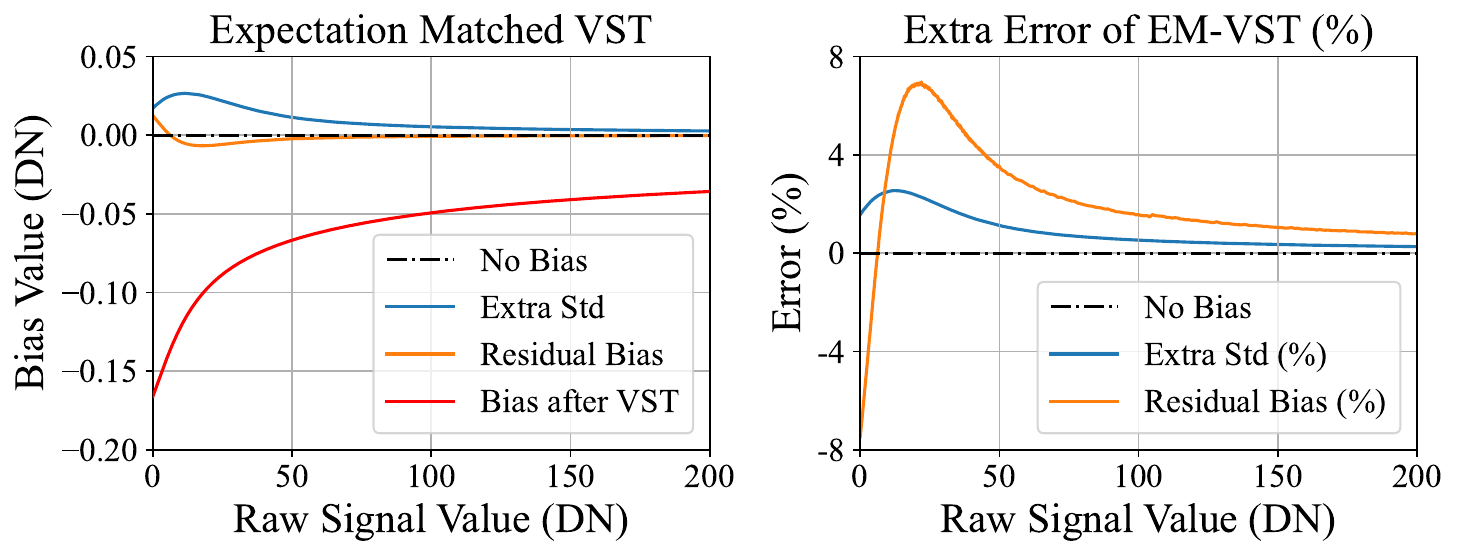}
		\caption{Analysis of additional errors introduced by EM-VST at ISO-3200 of a smartphone with an IMX686 sensor. The curves are generated by Monte Carlo sampling on synthetic noisy signals, where 10\(^6\) points are sampled for each raw signal value to estimate the bias relative to the true value.}
		\label{fig:error}
	\end{figure}
	
	We surprisingly discover a counter-intuitive solution in practice: expectation-matched VST. By using the noisy image \( y \) as the input to the bias function \( e_{\hat{\sigma}}(x) \), the bias can be pre-corrected with an approximate value \( \epsilon_y \). This approximate bias \( \epsilon_y \) can be derived via a Taylor expansion and formulated as
	\begin{equation}
		\epsilon_y = e_{\hat{\sigma}}(y) = e_{\hat{\sigma}}(x+n) = e_{\hat{\sigma}}(x) + e^{'}_{\hat{\sigma}}(x) \cdot n + \mathcal{O}(n^2),
	\end{equation}
	where \( n \) is the Poisson-Gaussian noise, \( e^{'}_{\hat{\sigma}}(x) \) is the first derivative of the bias function, and \( \mathcal{O}(n^2) \) represents the higher-order term of the Taylor expansion.
	
	Notably, the expectation of the Poisson-Gaussian noise \( n \) is zero. Furthermore, the first derivative \( e^{'}_{\hat{\sigma}}(x) \), being evaluated at a specific point \( x \), is treated as a constant in this context. As a result, the expectation of the first-order term \( e^{'}_{\hat{\sigma}}(x) \cdot n \) is also zero. Consequently, the expectation of the approximate bias \( \epsilon_y \) can be simplified to
	
	\begin{equation}
		\begin{aligned}\label{eq:errorE}
			E(\epsilon_y) &= E\left(e_{\hat{\sigma}}(x) + e^{'}_{\hat{\sigma}}(x) \cdot n + \mathcal{O}(n^2)\right) \\&= E(\epsilon_x) + 0 + E(\mathcal{O}(n^2)) \approx E(\epsilon_x).
		\end{aligned}
	\end{equation}
	
	As the higher-order term \( \mathcal{O}(n^2) \) is negligible, the expectation of first-order approximation \( \epsilon_y \) closely aligns with the expectation of exact bias \( \epsilon_x \), indicating that the bias is matched in expectation. The minimal homoscedastic noise introduced by EM-VST can be merged with the original image noise, allowing our flexible denoising model to effectively handle it.
	
	We have simulated and analyzed the behavior of EM-VST based on the calibrated noise parameters from various cameras. 
	The results reveal that the real-world bias function is generally smooth over its domain, with only slight gradient variations within the noise fluctuation range, as shown in Figure~\ref{fig:bias_3d}. 
	
	The presence of noise prevents the approximate bias $\epsilon_y$ from perfectly matching the exact bias $\epsilon_x$. Obtaining an analytical solution for their gap is challenging, we estimate the impact of the higher-order term \( \mathcal{O}(n^2) \) through numerical analysis.
	Typically, the additional standard deviation introduced by noise is within 3\% and the residual bias remains under 8\% as shown in Figure~\ref{fig:error}. After leaving the low-signal region, the values and proportions of additional noise and residual bias diminish progressively as signal levels increase.
	
	Based on the above analysis, directly computing the approximate bias \(\epsilon_y\) from the noisy image proves to be an efficient method for bias correction. Crucially, since the bias correction occurs before denoising, EM-VST neither amplifies denoising errors nor distorts image priors. 
	EM-VST exhibits low error in VST expectation bias correction, ensuring the exact color of denoised images.
	
	\begin{figure}[t]
		\centering
		\includegraphics[width=\linewidth, trim=5 5 5 0, clip]{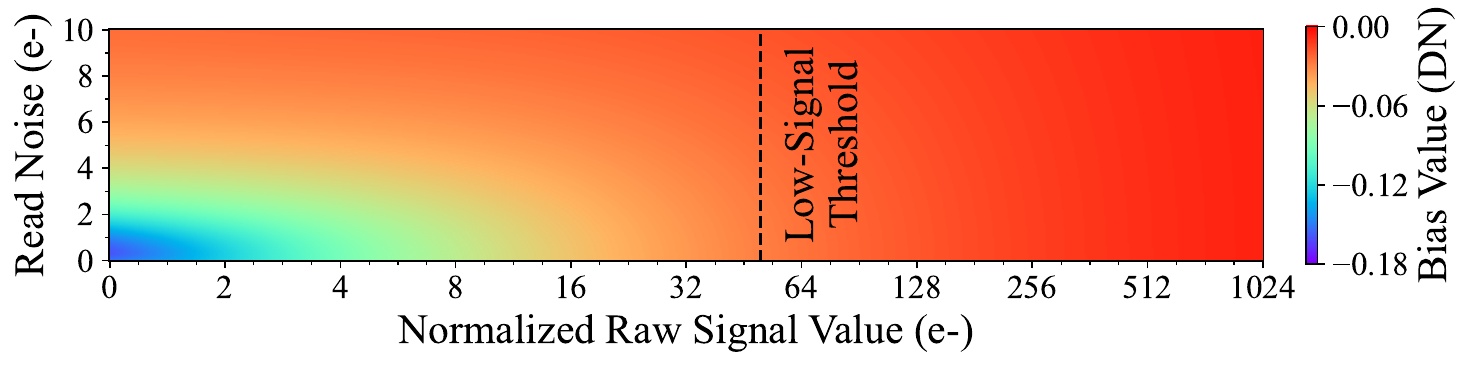}
		\caption{2D Look-Up Table (LUT) for bias correction. The normalized raw signal value (e-) represents the number of photoelectrons, while read noise (e-) indicates the noise intensity. The bias value represents the signal deviation in the raw image. The dashed line indicates the low-signal threshold ($x$=50). For optimal visualization, the horizontal axis employs a logarithmic scale.}

		\label{fig:lut}
	\end{figure}

	\subsubsection{\textbf{Procedure}}
	The procedure of EM-VST is straightforward. First, the bias function \(e_{\hat{\sigma}}(x)\) is fitted using the noise parameters \((\alpha, \sigma)\). Next, the noisy image \(y\) is mapped to the approximate bias \(\epsilon_y\) through the bias function \(e_{\hat{\sigma}}(x)\). Finally, the bias is corrected before denoising.  
	
	The primary challenge in implementing EM-VST lies in the computational complexity of the expectation term in the bias function, \ie, Eq.~\eqref{eq:intE}. To achieve a balance between precision and efficiency, we employ numerical integration in low-signal regions and a Taylor expansion-based closed-form solution~\cite{TIP11/O-VST,TIP13/VST} in high-signal regions.  
	Given the smooth and continuous nature of the bias function \(e_{\hat{\sigma}}(x)\), we further accelerate the computation by caching normalized results into a 2D look-up table (LUT), as shown in Figure~\ref{fig:lut}. With this strategy, EM-VST can process an image with 12M pixels in less than a second.  
	
	We observe that VST expectation bias sometimes disrupts image priors in low-signal regions. While such distortions are challenging to detect in common denoisers, they can significantly degrade the performance of generative denoising models.
	To support our claim, we conduct experiments on a diffusion-based denoiser~\cite{NIPS20/DDPM,ICLR21/DDIM,DMID}. We sample the noise parameter of SonyA7S2 at ISO-25600 with a digital gain of 300, corresponding to the darkest scenes in the SID dataset~\cite{CVPR18/SID}. According to the noise parameter, synthetic noisy raw images are generated to simulate extreme low-light conditions. We use GAT~\cite{GAT} with known noise parameters to transform the noisy images with Poisson-Gaussian noise into those with (approximately) AWGN. The transformed noisy images are adaptively embedded into the inference stage of a pre-trained diffusion model to obtain denoising results. 
	
	\begin{figure}[t]
		\footnotesize
		\setlength\tabcolsep{1pt}
		\centering
		\begin{tabular}{cccc}
			Noisy Image & GAT+UIAT~\cite{TIP13/VST} & EM-VST (Ours) & Reference \\
			{\includegraphics[width=0.23\linewidth]{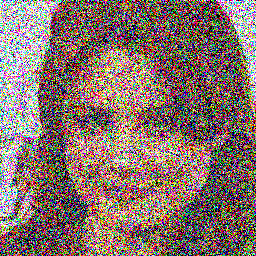}} &
			{\includegraphics[width=0.23\linewidth]{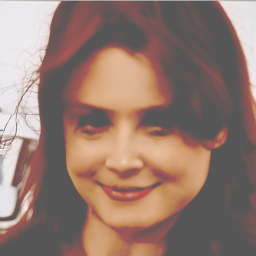}} &
			{\includegraphics[width=0.23\linewidth]{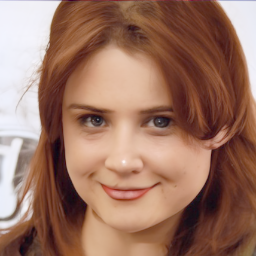}} & 
			{\includegraphics[width=0.23\linewidth]{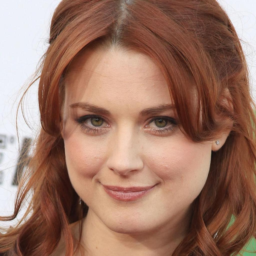}}
		\end{tabular}
		\caption{Comparison of different VST bias correction strategies in extreme low-light conditions based on a generative denoising method~\cite{DMID}.}
		\label{fig:DMID}
	\end{figure}
	
	In Figure~\ref{fig:DMID}, we compare the denoising results between the GAT+UIAT~\cite{TIP13/VST} and EM-VST. Since image priors are distorted before denoising, the method using GAT+UIAT remains incorrect color. Benefiting from the bias correction before denoising, the diffusion model with EM-VST successfully restores vivid images even in extreme low-light conditions. 
	
	It is worth noting that the diffusion model can also serve as an SNR-guided denoiser~\cite{DMID}, which inspires us to explore the diffusion extention of YOND.
	
	\subsection{SNR-guided Denoiser}\label{subsec:snr-net}
	A practical denoiser should be controllable to meet the diverse and complex challenges encountered in real-world scenarios. To fulfill the slogan of ``You Only Need a Denoiser", we propose an SNR-guided denoiser as a flexible and efficient solution.
	Theoretically, the noise variance of the transformed noisy image is approximately stabilized at 1 after VST. The SNR of noisy images can be easily obtained by calculating the ratio of their peak value to 1. The reciprocal of the SNR corresponds to the noise level $\sigma_{\text{SNR}}$, which serves as explicit guidance for AWGN denoising. Under adaptive adjustment of this guidance value, SNR-Net can precisely control the denoising process, delivering clear denoised images across diverse challenging real-world scenarios.
	
	\subsubsection{\textbf{Structure}}
\begin{figure}[t]
		\centering
		\includegraphics[width=\linewidth]{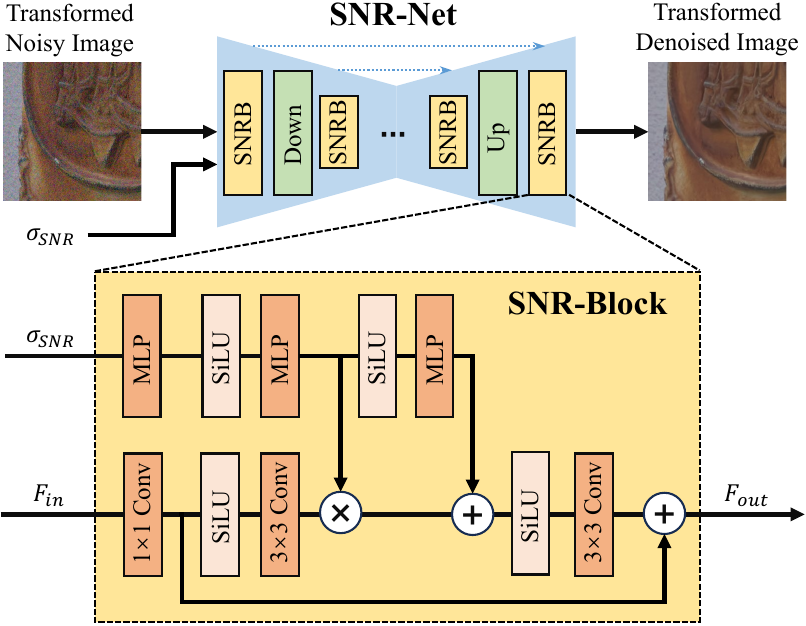}
		\caption{Detailed network structure of our SNR-Net. $\sigma_{\text{SNR}}$ indicates the noise level of the transformed noisy image provided by CNE. $F_{in}$ and $F_{out}$ represent the input features and output features, respectively.}
		\label{fig:SNR-Net}
	\end{figure}

	To highlight that the strength of YOND lies primarily in its pipeline rather than in the specific neural network architecture, we apply minimal yet essential modifications to a UNet~\cite{Unet} backbone with 32 base channels. These straightforward adaptations yield a simple but effective SNR-guided denoiser named SNR-Net, as shown in Figure~\ref{fig:SNR-Net}. SNR-Net consists of four downsampling and four upsampling stages. Downsampling is performed using 3$\times$3 convolutions with a stride of 2 while upsampling is achieved with 2$\times$2 transposed convolutions with a stride of 2. The number of channels doubles at each downsampling stage. Each level contains an SNR-Block. The SNR-Block is a residual block that incorporates an SNR-guided branch. The noise level $\sigma_{\text{SNR}}$ is provided as an input to each SNR-Block, guiding the network to learn controllable denoising. We adopt SiLU~\cite{SiLU} as our activation function. The input $\sigma_{\text{SNR}}$ can be flexibly adjusted to manually explore optimal image quality in practice. For instance, amplifying $\sigma_{\text{SNR}}$ by 3\% effectively counteracts the additional noise introduced by EM-VST.
	
	In summary, SNR-Net offers controllable raw AWGN denoising, enabling YOND to deliver clear denoised images under adaptive adjustments.
	
	\subsubsection{\textbf{Diffusion Extension}}\label{sec:YOND-p}
	We adhere to the principle that ``form follows function"~\cite{sullivan1922tall}, indicating that similar forms can support similar functions. The diffusion model and the denoising task share structural similarities, prompting us to harness the deeper potential of SNR-Net with advanced diffusion models. DMID~\cite{DMID} has demonstrated that a powerful diffusion model can also serve as an SNR-guided denoiser, with its pre-trained model directly applicable to generative denoising. Building on this insight, we propose an iterative denoising strategy based on DDIM~\cite{ICLR21/DDIM} to extend SNR-Net into a simplified diffusion model. We name the YOND operating under this strategy as YOND-p, which offers the possibility to further improve image perceptual quality.
	
	The proposed method is summarized in Algorithm~\ref{alg:ddim}. We assume a total of $T$ iterations, all performed after the VST. The initial denoising step follows the standard YOND process, where the SNR-Net denoises the noisy image $x_T$ guided by the noise level $\sigma_{\text{SNR}}$. After obtaining the initial denoising result, the generative iterative denoising process begins. First, the predicted noise $\epsilon_t$ is updated based on the original noisy image $x_T$ and the denoised image $x_0$. Then, following the DDIM strategy, the predicted noise is mixed with random AWGN, with the mixing factor denoted as $\eta$. Finally, the noisy image $x_t$ and noise level $\sigma_t$ are updated, and the denoised image $x_0$ is further refined. The final denoised image $x_0$ from the last iteration is considered the output.
	
	\begin{algorithm}[t]
		\caption{DDIM-based Iterative Denoising Strategy}\label{alg:ddim}
		\begin{algorithmic}[1]
			\State \textbf{Input:} Noisy image $x_T$, Pretrained SNR-Net $f_{\theta}$, Initial noise level $\sigma_{\text{SNR}}$, Iteration count $T$, Noise mix factor $\eta$, Noise decay ratio $\gamma$
			\State $x_0 \gets f_{\theta}(x_T, \sigma_{\text{SNR}})$ \Comment{Initial denoising}
			\For {$t=T-1$ to $1$}
			\State $\epsilon_t \gets \gamma^{T-t-1} (x_T - x_0)$ \Comment{Update predicted noise $\epsilon_t$}
			\State Sample $z \sim \mathcal{N}(0,1)$
			\State $\epsilon_t \gets \eta \epsilon_t + \sqrt{1 - \eta^2} \sigma_t z$ \Comment{DDIM noise mix-up}
			\State $x_t \gets x_0 + \gamma \epsilon_t$ \Comment{Update noisy image $x_t$}
			\State $\sigma_t \gets \gamma^{T-t}\sigma_{\text{SNR}}$ \Comment{Update noise level $\sigma_t$}
			\State $x_0 \gets f_{\theta}(x_t, \sigma_t)$ \Comment{Update denoised image $x_0$}
			\EndFor
			\State \textbf{Output:} Denoised image $x_0$
		\end{algorithmic}
	\end{algorithm}
	
	There are some details deserving emphasis in our iterative denoising strategy. 
	First, it is crucial to update the noise $\epsilon_t$ based on the original noisy image $x_T$ rather than the noisy image from the previous step. This design prevents the true details in $x_T$ from being excessively suppressed and permanently lost during the early stages of iteration.
	Second, since the SNR-Net has significantly lower computational complexity compared to a standard diffusion model, the noise mix factor $\eta$ must be close to 1 to reduce the computational burden of generating details, a consideration also reported in some diffusion model distillation studies~\cite{CVPR24/SinSR}. 
	Lastly, we find that the SNR-Net with 32 base channels lacks sufficient generative capability, resulting in performance akin to simple sharpening. Consequently, we double the number of base channels and train a new SNR-Net to serve as the denoiser of YOND-p.
	
	YOND-p focuses on perceptual quality rather than distortion recovery, placing it on a different track from classical denoising methods. Therefore, YOND-p is not included in the main experiments in Section~\ref{sec:exp}. We will provide specific use cases of YOND-p in Section~\ref{sec:discuss_YOND-p}.
	
	\begin{table}[t]
		\footnotesize
		\setlength\tabcolsep{4pt}
		\caption{The computational complexity comparison of denoising networks mentioned in this paper}\label{tab:cost}
		\centering
			\begin{tabular}{lcrr}
				\toprule
				\makebox[0.240\linewidth][l]{Method} & \makebox[0.130\linewidth][c]{Structure} & \makebox[0.110\linewidth][c]{Parameters} & \makebox[0.090\linewidth][c]{Macs} \\
				\midrule
				\rowcolor{lightgray}
				\makecell[l]{N2C, N2N~\cite{ICML18/N2N}, NBR2NBR~\cite{CVPR21/NBR2NBR}\\N2V~\cite{CVPR19/N2V}, B2U~\cite{CVPR22/B2U}, DCD~\cite{ICCV23/DCD}} & UNet~\cite{Unet} & 1.1 M & 4.71 G \\
				FBI~\cite{CVPR21/FBI} & FBI-Net~\cite{CVPR21/FBI} & 0.81 M & 53.47 G \\
				\rowcolor{lightgray}
				P-G~\cite{P-G}, ELD~\cite{TPAMI21/ELD}, SFRN~\cite{ICCV21/SFRN} & UNet~\cite{Unet} & 7.76 M & 3.44 G \\
				Diffusion Models~\cite{NIPS20/DDPM, ICLR21/DDIM,DMID} & UNet~\cite{Unet} & 552.81 M & 278.67 G \\
				\midrule
				\rowcolor{lightgray}
				YOND & SNR-Net & 11.17 M & 3.71 G \\
				YOND-p & SNR-Net & 44.67 M & 14.83 G \\
				\bottomrule
			\end{tabular}
	\end{table}
	
	\subsubsection{\textbf{Computational Complexity}}
	
	Table~\ref{tab:cost} presents a comparison of the computational complexity of various methods and network structures discussed in this paper. The complexity is calculated using the open-source package \textit{ptflops} on a 256$\times$256 raw image. Our method achieves high practicality at a moderate computational complexity.
	
	\begin{table*}[t]
		\footnotesize
		\setlength\tabcolsep{7pt}
		\caption{Quantitative results (PSNR/SSIM) of calibration-base methods and YOND on the ELD datasets~\cite{TPAMI21/ELD} and LRID dataset~\cite{TPAMI23/PMN}}\label{tab:nm}
		\centering
			\begin{tabular}{c|c|c|cc|cccc}
				\Xhline{0.8pt}\rule{0pt}{8pt}
				\makebox[0.040\textwidth][c]{}  & \makebox[0.040\textwidth][c]{} & \makebox[0.040\textwidth][c]{} &
				\multicolumn{2}{c}{w/ Camera-specific Calibration}\vline &
				\multicolumn{4}{c}{w/o Camera-specific Calibration}
				\\
				\Xcline{4-9}{0.4pt}\rule{0pt}{8pt}
				\multirow{-2.2}{*}{\makecell[c]{Dataset}} & \multirow{-2.2}{*}{\makecell[c]{Subset}} & \multirow{-2.2}{*}{\makecell[c]{ISO ($\times$ratio)}} & \makebox[0.080\textwidth][c]{P-G~\cite{P-G}} &
				\makebox[0.080\textwidth][c]{ELD~\cite{TPAMI21/ELD}}&
				\makebox[0.080\textwidth][c]{P-G (Blind)} & 
				\makebox[0.080\textwidth][c]{ELD (Blind)} & 
				\makebox[0.080\textwidth][c]{YOND (UNet)} & 
				\makebox[0.080\textwidth][c]{YOND} \\
				\Xhline{0.6pt}\rule{0pt}{8pt}
				\multirow{4.5}{*}{\makecell[c]{ELD}} & \multirow{2}{*}{\makecell[c]{SonyA7S2}} & [800,3200] & {54.94} / {{0.998}} & {53.29} / {{0.996}} & {53.22} / {{0.998}} & {52.67} / {{0.997}} & \textbf{\color{blue}56.39} / {\textbf{\color{red}0.999}} & \textbf{\color{red}57.98} / {\textbf{\color{red}0.999}} \\
				{} & {} & [8000,32000] & {50.52} / {{0.990}} & {50.45} / {\textbf{\color{blue}0.992}} & {50.81} / {\textbf{\color{blue}0.992}} & {50.76} / {\textbf{\color{red}0.993}} & \textbf{\color{blue}51.87} / {\textbf{\color{blue}0.992}} & \textbf{\color{red}52.66} / {\textbf{\color{red}0.993}} \\
				\Xcline{2-9}{0.4pt}\rule{0pt}{8pt}
				{} & \multirow{2}{*}{\makecell[c]{NikonD850}} & [800,3200] & {50.73} / {{0.995}} & {50.61} / {{0.992}} & {51.83} / {{0.995}} & {49.18} / {{0.985}} & \textbf{\color{blue}53.42} / {\textbf{\color{blue}0.997}} & \textbf{\color{red}54.96} / {\textbf{\color{red}0.998}} \\
				{} & {} & [8000,32000] & {48.46} / {\textbf{\color{blue}0.989}} & {48.42} / {\textbf{\color{blue}0.989}} & {48.00} / {{0.987}} & {46.68} / {{0.980}} & \textbf{\color{blue}48.68} / {\textbf{\color{blue}0.989}} & \textbf{\color{red}49.67} / {\textbf{\color{red}0.991}} \\
				\Xhline{0.6pt}\rule{0pt}{8pt}
				\multirow{4.5}{*}{\makecell[c]{LRID}} & \multirow{2}{*}{\makecell[c]{Indoor}} & 6400 & {47.92} / {\textbf{\color{red}0.990}} & {48.13} / {\textbf{\color{red}0.990}} & {47.74} / {\textbf{\color{blue}0.989}} & {47.75} / {{0.988}} & \textbf{\color{blue}48.39} / {\textbf{\color{red}0.990}} & \textbf{\color{red}48.99} / {\textbf{\color{red}0.990}} \\
				{} & {} & 12800 & {46.03} / \textbf{\color{blue}{0.982}} & \textbf{\color{blue}46.48} / {\textbf{\color{red}0.983}} & {46.21} / {\textbf{\color{red}0.983}} & {46.13} / {\textbf{\color{red}0.983}} & {46.18} / {{0.980}} & \textbf{\color{red}46.72} / {{0.981}} \\
				\Xcline{2-9}{0.4pt}\rule{0pt}{8pt}
				{} & \multirow{2}{*}{\makecell[c]{Outdoor}} & 6400 & \textbf{\color{blue}45.27} / {\textbf{\color{blue}0.985}} & {45.24} / {\textbf{\color{blue}0.985}} & {45.02} / {{0.984}} & {44.82} / {{0.983}} & {45.20} / {\textbf{\color{blue}0.985}} & \textbf{\color{red}46.07} / {\textbf{\color{red}0.987}} \\
				{} & {} & 12800 & \textbf{\color{blue}43.76} / {\textbf{\color{red}0.975}} & {43.68} / {\textbf{\color{blue}0.974}} & {43.34} / {\textbf{\color{blue}0.974}} & {43.14} / {\textbf{\color{blue}0.974}} & {43.48} / {{0.971}} & \textbf{\color{red}44.09} / {{0.972}} \\
				\Xhline{0.8pt}
			\end{tabular}
			\begin{flushleft}
				$^1$ The \textbf{\color{red}red} scores denote the best results and the \textbf{\color{blue}blue} scores denote the second-best results. \\
				$^2$ For calibration-based methods, the denoiser of ``Blind" is trained by the noise parameters calibrated on another camera. 
			\end{flushleft}
	\end{table*}
	
	\begin{figure*}[!t]
		\small
		\setlength\tabcolsep{2pt}
		\renewcommand\arraystretch{0.8}
		\begin{center}
			\begin{tabular}{@{} c c @{}}
				\begin{tabular}{@{} c @{}}
					\includegraphics[width=.4410\linewidth]{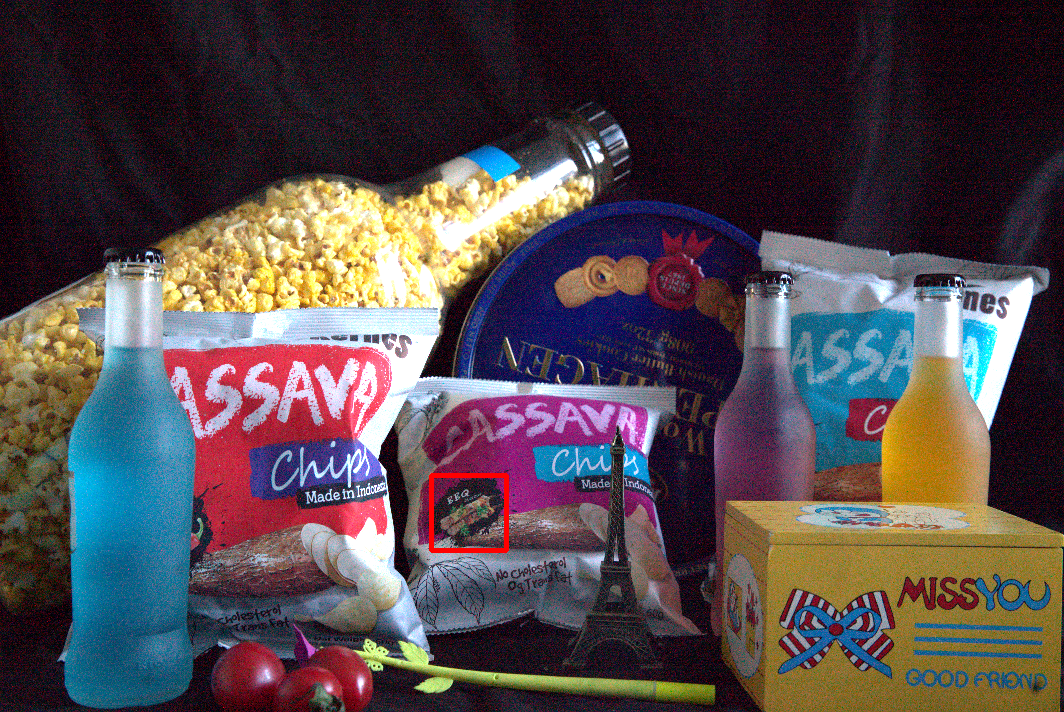} \\
					ELD-SonyA7S2 \\
					Scene-03 / IMG-0008 \\
					\\
					\includegraphics[width=.4410\linewidth]{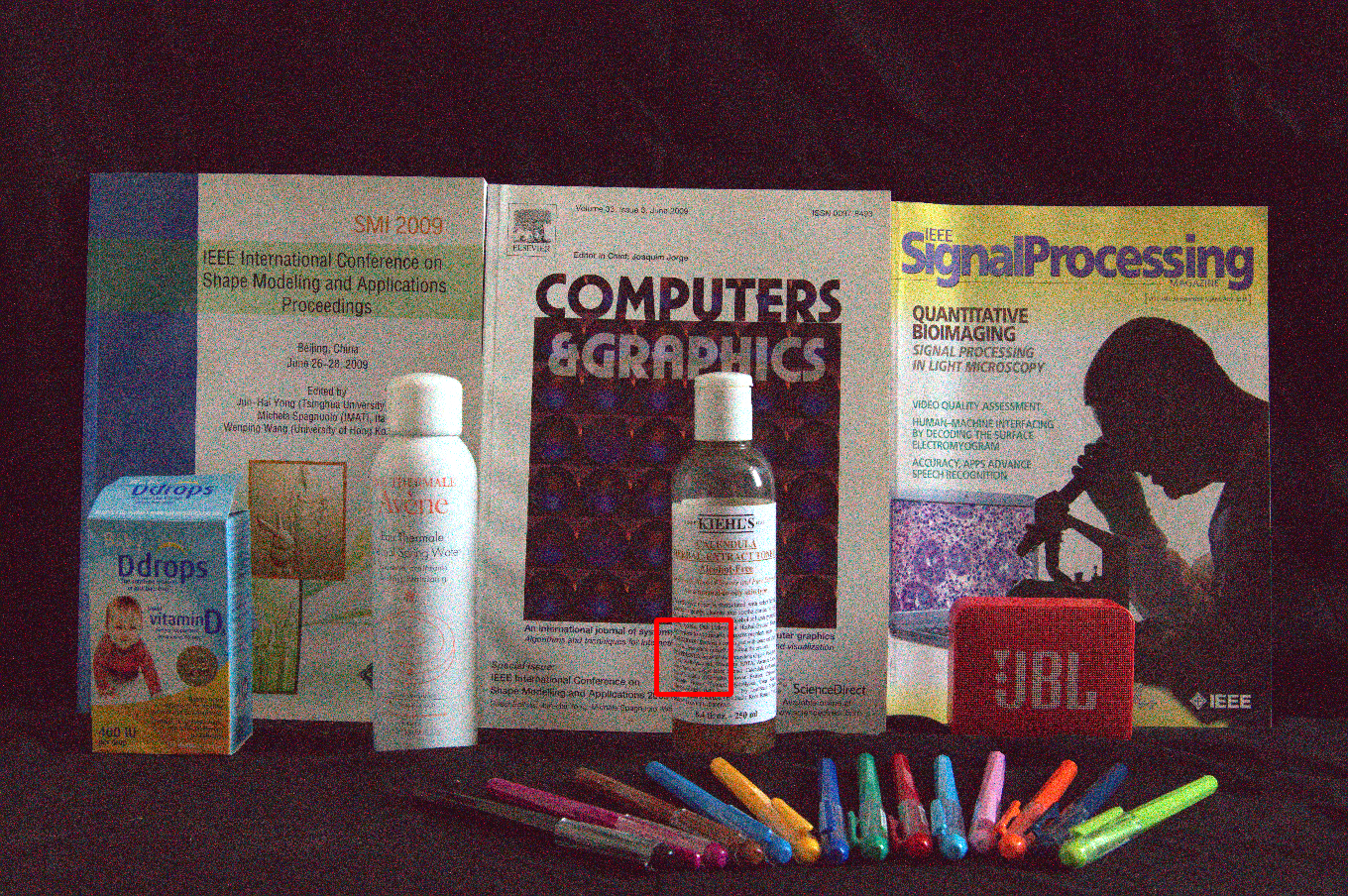} \\
					ELD-NikonD850 \\
					Scene-06 / IMG-0013 \\
				\end{tabular} & 
				\begin{tabular}{@{} c c c c @{}}
					\includegraphics[width=.1280\linewidth]{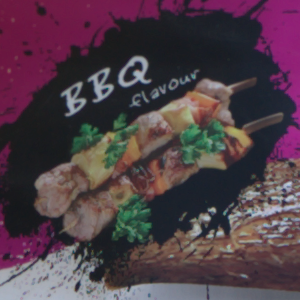} & 
					\includegraphics[width=.1280\linewidth]{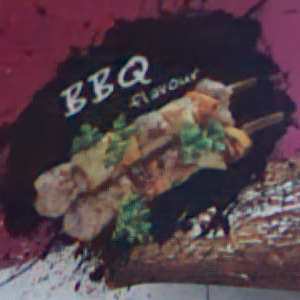} &
					\includegraphics[width=.1280\linewidth]{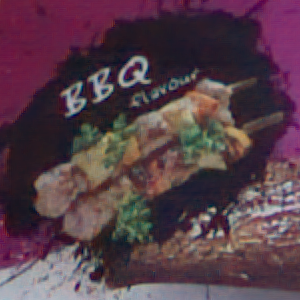} &
					\includegraphics[width=.1280\linewidth]{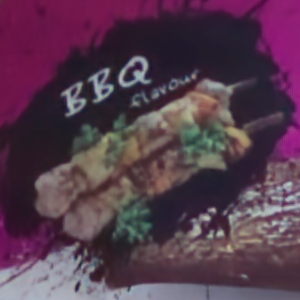} \\
					Reference& P-G (Blind) & ELD (Blind) & YOND (UNet)\\
					PSNR / SSIM & {44.79} / {0.9862} & {42.73} / {0.9797} & \textbf{\color{blue}46.34} / {0.9904}\\
					\includegraphics[width=.1280\linewidth]{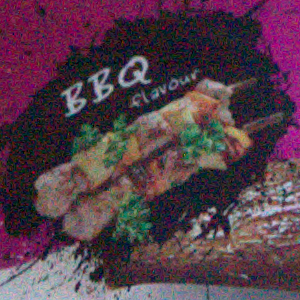} &
					\includegraphics[width=.1280\linewidth]{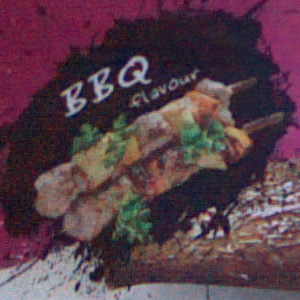} &
					\includegraphics[width=.1280\linewidth]{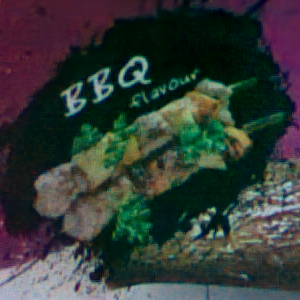} &
					\includegraphics[width=.1280\linewidth]{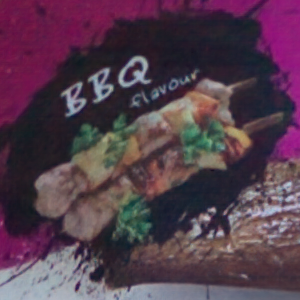} \\
					Noisy Image & P-G\cite{P-G} & ELD\cite{TPAMI21/ELD} &  YOND\\
					{40.64} / {0.9436} & {44.60} / {0.9898} & {43.86} / \textbf{\color{blue}0.9907} & \textbf{\color{red}47.67} / \textbf{\color{red}0.9922}\\
					
					\\
					\includegraphics[width=.1280\linewidth]{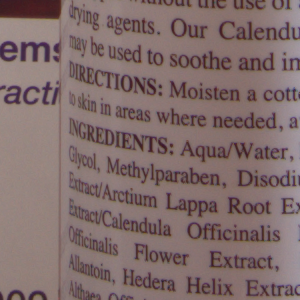} & 
					\includegraphics[width=.1280\linewidth]{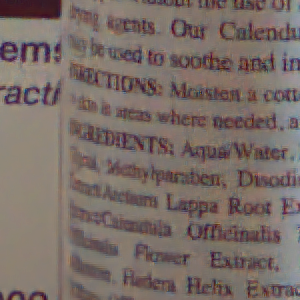} &
					\includegraphics[width=.1280\linewidth]{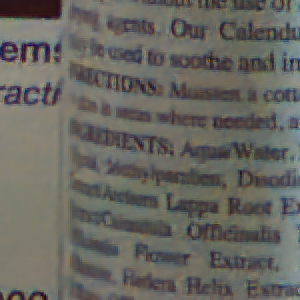} &
					\includegraphics[width=.1280\linewidth]{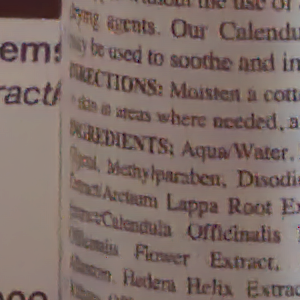} \\
					Reference& P-G (Blind) & ELD (Blind) & YOND (UNet) \\
					PSNR / SSIM & {43.44} / {0.9806} & {39.43} / {0.9686} & \textbf{\color{blue}44.77} / \textbf{\color{blue}0.9856}\\
					\includegraphics[width=.1280\linewidth]{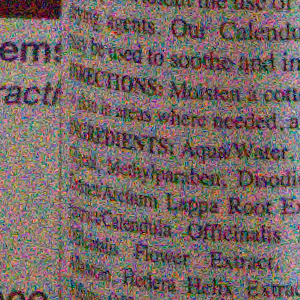} &
					\includegraphics[width=.1280\linewidth]{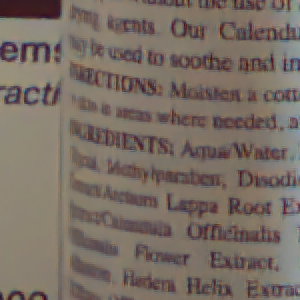} &
					\includegraphics[width=.1280\linewidth]{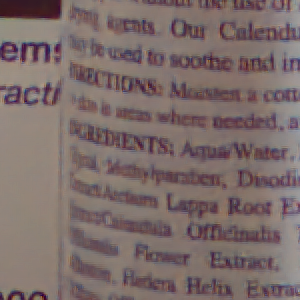} &
					\includegraphics[width=.1280\linewidth]{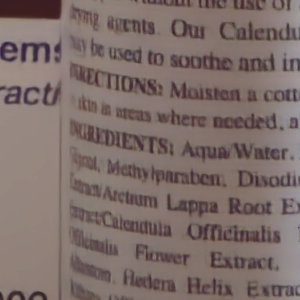} \\
					Noisy Image & P-G\cite{P-G} & ELD\cite{TPAMI21/ELD} &  YOND\\
					{32.46} / {0.7418} & {43.39} / {0.9818} & {43.47} / {0.9819} & \textbf{\color{red}45.79} / \textbf{\color{red}0.9887}\\
					
				\end{tabular}
			\end{tabular}
		\end{center}
		\caption{Raw image denoising results on images from the ELD dataset. The \textbf{\color{red}red} color indicates the best results and the \textbf{\color{blue}blue} color indicates the second-best results. \textbf{(Best viewed with zoom-in)}}
		\label{fig:ELD-compare}
	\end{figure*}
	
	\begin{figure*}[!t]
		\small
		\setlength\tabcolsep{2pt}
		\renewcommand\arraystretch{0.8}
		\begin{center}
			\begin{tabular}{@{} c c @{}}
				\begin{tabular}{@{} c @{}}
					\includegraphics[width=.3920\linewidth]{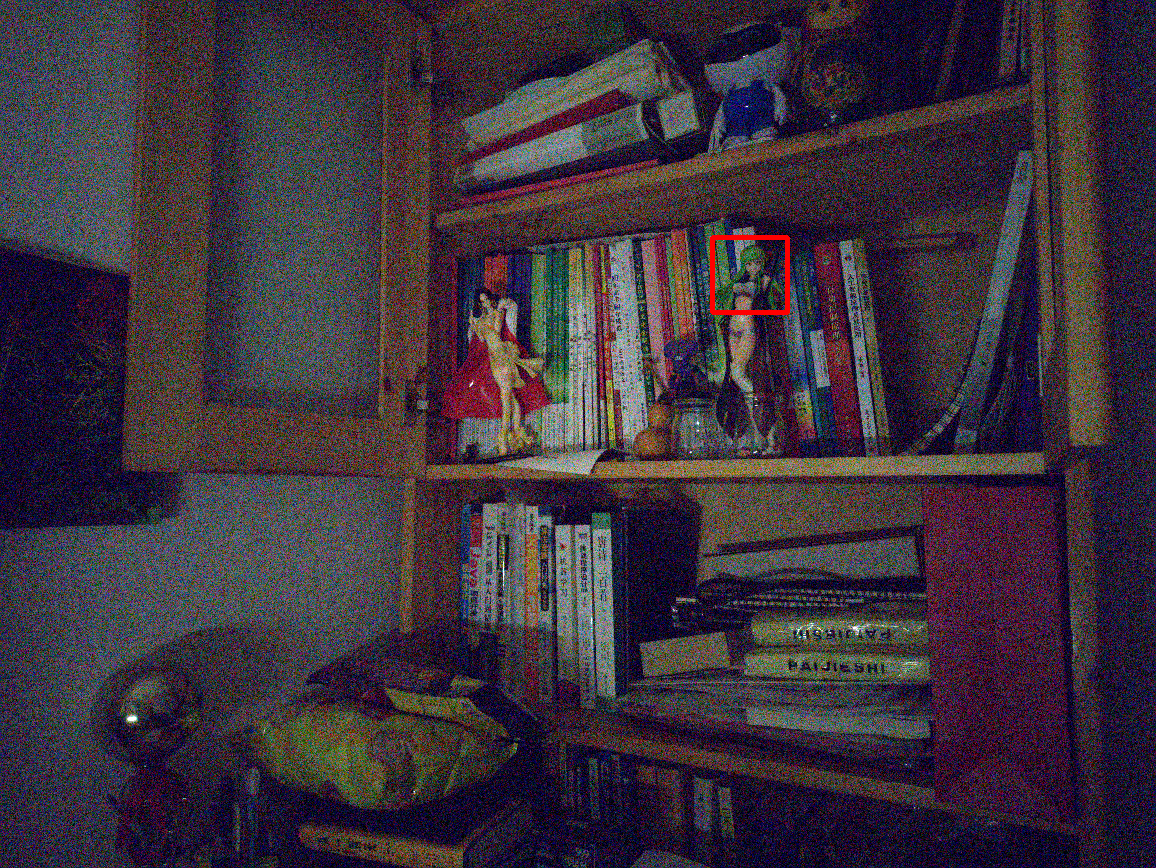} \\
					LRID-Indoor \\
					Scene-004 \\
					\\
					\includegraphics[width=.3920\linewidth]{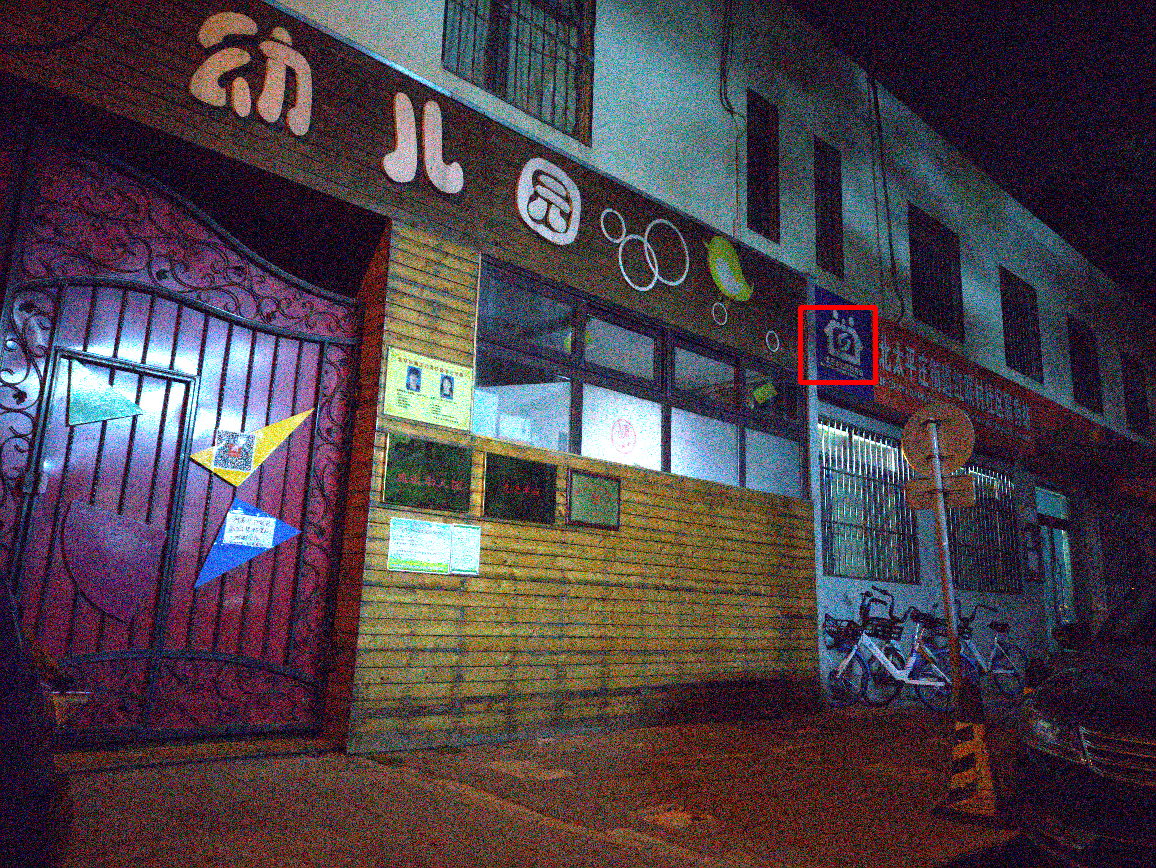} \\
					LRID-Outdoor \\
					Scene-022 \\
				\end{tabular} & 
				\begin{tabular}{@{} c c c c @{}}
					\includegraphics[width=.1280\linewidth]{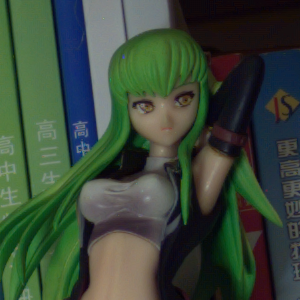} & 
					\includegraphics[width=.1280\linewidth]{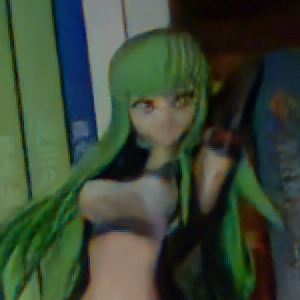} &
					\includegraphics[width=.1280\linewidth]{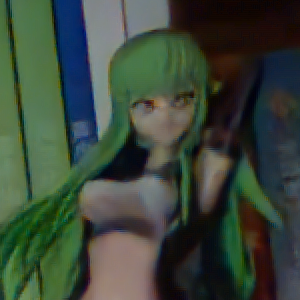} &
					\includegraphics[width=.1280\linewidth]{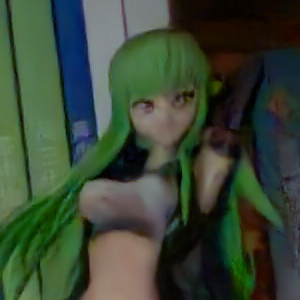} \\
					Reference& P-G (Blind) & ELD (Blind) & YOND (UNet)\\
					PSNR / SSIM & {48.75} / {0.9907} & {48.75} / \textbf{\color{blue}0.9915} & {49.10} / {0.9902} \\
					\includegraphics[width=.1280\linewidth]{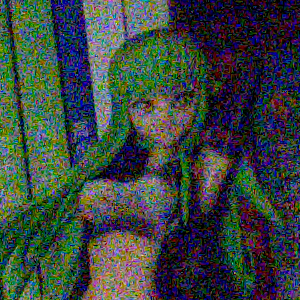} &
					\includegraphics[width=.1280\linewidth]{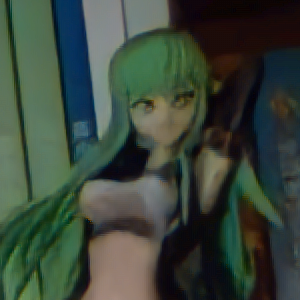} &
					\includegraphics[width=.1280\linewidth]{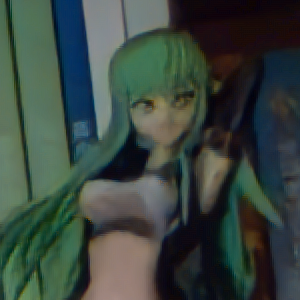} &
					\includegraphics[width=.1280\linewidth]{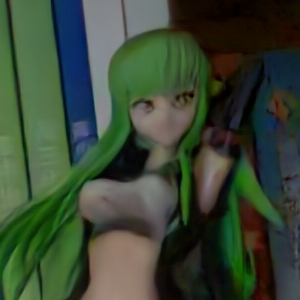} \\
					Noisy Image & P-G\cite{P-G} & ELD\cite{TPAMI21/ELD} &  YOND\\
					{33.77} / {0.7186} & {48.91} / {0.9912} & \textbf{\color{blue}49.13} / \textbf{\color{red}0.9921} & \textbf{\color{red}49.80} / {0.9910}\\
					
					\\
					\includegraphics[width=.1280\linewidth]{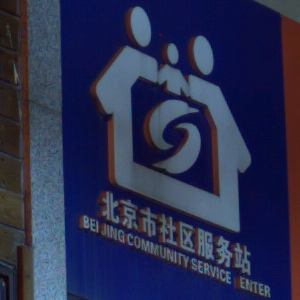} & 
					\includegraphics[width=.1280\linewidth]{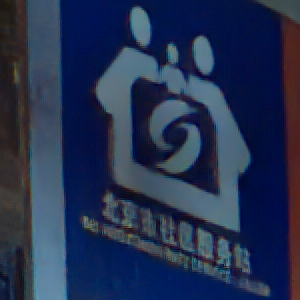} &
					\includegraphics[width=.1280\linewidth]{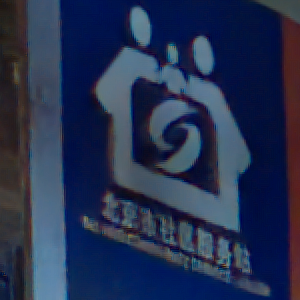} &
					\includegraphics[width=.1280\linewidth]{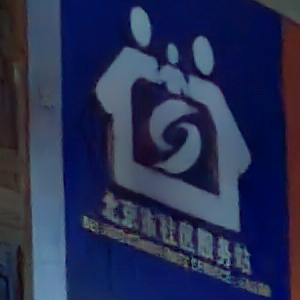} \\
					Reference& P-G (Blind) & ELD (Blind) & YOND (UNet) \\
					PSNR / SSIM & {41.95} / {0.9812} & {41.80} / {0.9807} & \textbf{\color{blue}42.92} / \textbf{\color{blue}0.9836} \\
					\includegraphics[width=.1280\linewidth]{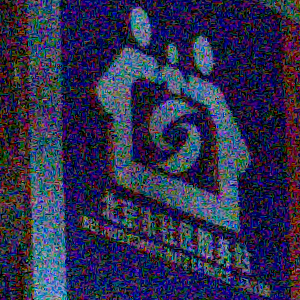} &
					\includegraphics[width=.1280\linewidth]{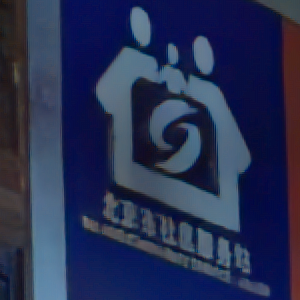} &
					\includegraphics[width=.1280\linewidth]{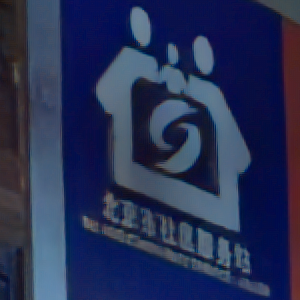} &
					\includegraphics[width=.1280\linewidth]{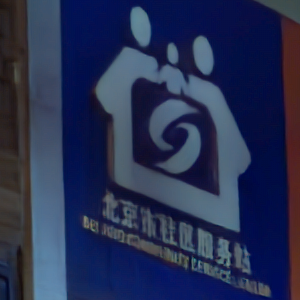} \\
					Noisy Image & P-G\cite{P-G} & ELD\cite{TPAMI21/ELD} &  YOND\\
					{31.94} / {0.6982} & {42.15} / {0.9820} & {42.08} / {0.9791} & \textbf{\color{red}43.48} / \textbf{\color{red}0.9853}\\
					
				\end{tabular}
			\end{tabular}
		\end{center}
		\caption{Raw image denoising results on images from the LRID dataset. The \textbf{\color{red}red} color indicates the best results and the \textbf{\color{blue}blue} color indicates the second-best results. \textbf{(Best viewed with zoom-in)}}
		\label{fig:IMX686-compare}
	\end{figure*}
	
	\section{Experiments}\label{sec:exp}
	\subsection{Data and Experimental Settings}
	\subsubsection{\textbf{Training Details of SNR-Net}}
	We regard the DIV2K dataset~\cite{CVPR17/DIV2K} and SID dataset~\cite{CVPR18/SID} as training data source. The training set of the DIV2K dataset and SID dataset includes 800 high-resolution clean images and 161 high-bit clean images (processed from raw images), respectively. We crop them into 56294 non-overlapping 256$\times$256 patches. We adopt the unprocessing~\cite{CVPR19/Unprocess} technique to convert sRGB images into pseudo raw images with different Bayer patterns~\cite{CVPRW19/fhq}. The AWGN denoiser is trained with normalized synthetic noisy-clean image pairs. According to the noise parameter range of multiple calibrated cameras, we add AWGN noise to clean pseudo raw image with noise level $\sigma$ $\in$ [5,50]. This noise level range has covered the majority of scenes within ISO 20000 for most smartphones. We train denoising models with 600 epochs using Adam optimizer~\cite{Adam} and $\mathcal L_1$ loss. The learning rate will vary with each epoch in a cosine annealing manner~\cite{ICLR17/SGDR}. The base learning rate is set to 2$\times$10$^{-4}$ and the minimum learning rate is set to 10$^{-5}$. The optimizer restarts every 200 epochs and the learning rate is halved on restarts.
	
	\subsubsection{\textbf{Evaluation Setting}}
	We focus on the practicality of YOND, thus all main experiments are conducted on real-world raw image denoising datasets. YOND breaks down the dataset barriers in the raw denoising domain, enabling direct comparisons with self-supervised methods and calibration-based methods in their respective domains of expertise.
	
	For comparisons with calibration-based methods, we conduct experiments on the ELD dataset~\cite{TPAMI21/ELD} and LRID dataset~\cite{TPAMI23/PMN}. Both datasets are designed for noise modeling and low-light raw image denoising, with some low-light data having noise level far beyond the capacity of general denoisers. We experiment on subsets of the validation set with noise level $\sigma\textless$50. Due to the difficulty on calibration materials collecting, we just select P-G~\cite{P-G} and ELD~\cite{TPAMI21/ELD} for comparison, whose model weights are provided by LED~\cite{ICCV23/LED} and PMN~\cite{TPAMI23/PMN}. Since the compared methods are both based on the same UNet, we provide the results of YOND (UNet) with the same network structure for reference.
	
	For comparisons with self-supervised methods, we conduct experiments on the LRID dataset, SIDD dataset~\cite{CVPR18/SIDD}, and DND dataset~\cite{CVPR17/DND}. The SIDD dataset includes 10 scenes captured with 5 smartphones under diverse settings. The DND dataset covers 50 scenes captured with 4 different cameras under diverse settings. Due to the lack of available code and reproducibility for many self-supervised methods, we focus on recent open-source projects for comparison. We will compare our method with VST+BM3D~\cite{TIP13/VST,BM3D}, N2N~\cite{ICML18/N2N},  N2V~\cite{CVPR19/N2V}, FBI-Net~\cite{CVPR21/FBI}, NBR2NBR~\cite{CVPR21/NBR2NBR}, B2U~\cite{CVPR22/B2U}, and DCD~\cite{ICCV23/DCD}, where the pre-trained weights for N2N, N2V, and DCD are kindly supplied by the authors of DCD.
	Notably, VST+BM3D with single-image noise estimation is a powerful traditional blind denoising method, which serves as our baseline. We equip VST+BM3D with an optimized single-image noise estimation method~\cite{P-G} for a fair comparison.
	
	It is worth noting that \textbf{YOND only needs to be trained once on synthetic data}. The performances of YOND are evaluated under the same inference pipeline and weights across all datasets. YOND performs well despite these inherent data limitations, demonstrating its superior practicality.
	
	\begin{figure*}[t!]
		\small 
		\setlength\tabcolsep{1pt}
		\renewcommand\arraystretch{0.8}
		\centering
		\begin{center}
			{%
				\begin{tabular}{cccccccc}
					\makebox[0.120\textwidth][c]{Input} & 
					\makebox[0.120\textwidth][c]{VST+BM3D~\cite{BM3D}} &
					\makebox[0.120\textwidth][c]{NBR2NBR~\cite{CVPR21/NBR2NBR}} & 
					\makebox[0.120\textwidth][c]{FBI~\cite{CVPR21/FBI}} & 
					\makebox[0.120\textwidth][c]{B2U~\cite{CVPR22/B2U}} &
					\makebox[0.120\textwidth][c]{DCD~\cite{ICCV23/DCD}} &  
					\makebox[0.120\textwidth][c]{YOND} & 
					\makebox[0.120\textwidth][c]{Reference} \\
					{\includegraphics[width=0.120\textwidth]{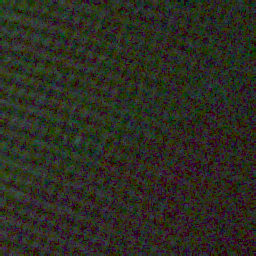}}& 
					{\includegraphics[width=0.120\textwidth]{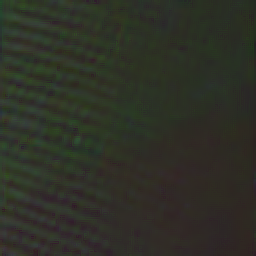}} & 
					{\includegraphics[width=0.120\textwidth]{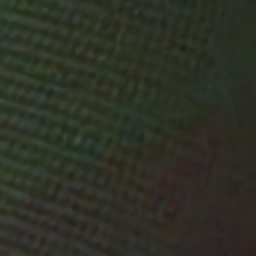}}&
					{\includegraphics[width=0.120\textwidth]{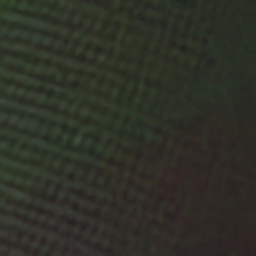}}&
					{\includegraphics[width=0.120\textwidth]{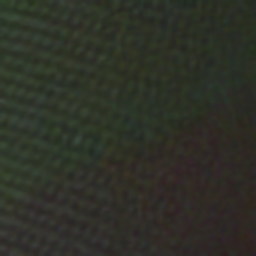}}&
					{\includegraphics[width=0.120\textwidth]{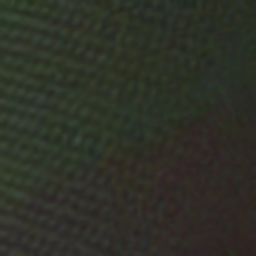}} &
					{\includegraphics[width=0.120\textwidth]{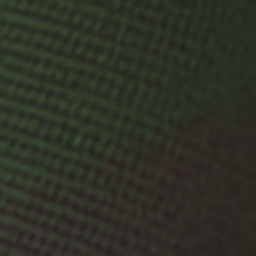}}&
					{\includegraphics[width=0.120\textwidth]{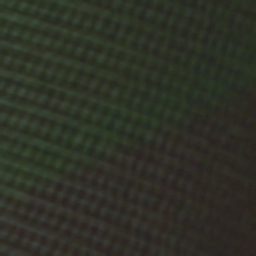}} \\
					39.89 / 0.9077 & 50.43 / 0.9911 & 51.51 / 0.9930 & \textbf{\color{blue}{52.33}} / \textbf{\color{blue}{0.9941}} & 51.91 / 0.9935 & 51.96 / 0.9936 & \textbf{\color{red}{52.89}} / \textbf{\color{red}{0.9949}} & PSNR / SSIM \\
					{\includegraphics[width=0.120\textwidth]{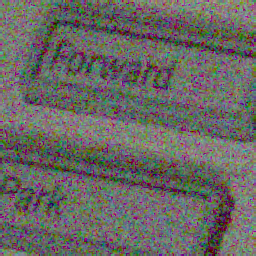}}& 
					{\includegraphics[width=0.120\textwidth]{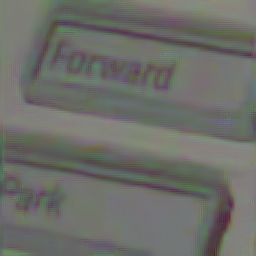}} & 
					{\includegraphics[width=0.120\textwidth]{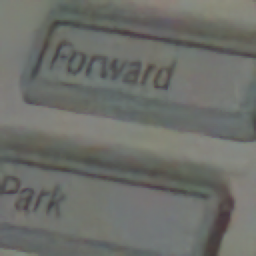}}&
					{\includegraphics[width=0.120\textwidth]{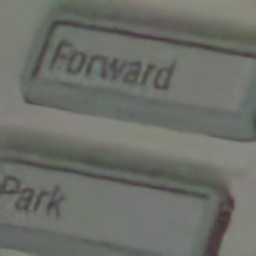}}&
					{\includegraphics[width=0.120\textwidth]{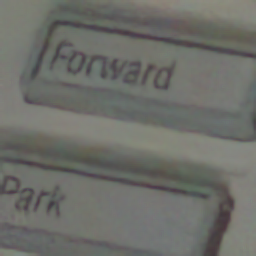}}&
					{\includegraphics[width=0.120\textwidth]{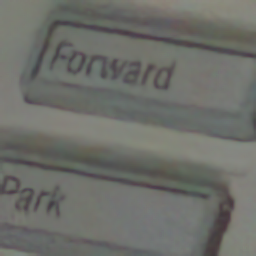}} &
					{\includegraphics[width=0.120\textwidth]{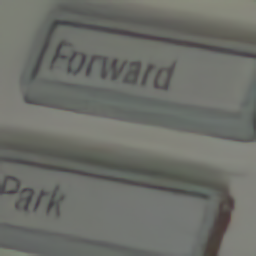}}&
					{\includegraphics[width=0.120\textwidth]{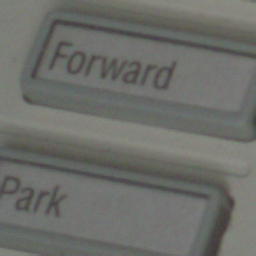}} \\
					26.23 / 0.6168 & 39.13 / 0.9711 & 40.69 / 0.9798 & 41.17 / 0.9826 & {41.55} / {0.9839} & \textbf{\color{blue}41.68} / \textbf{\color{blue}0.9844} & \textbf{\color{red}42.32} / \textbf{\color{red}0.9867} & PSNR / SSIM \\
					\addlinespace[1.2pt]
					{\includegraphics[width=0.120\textwidth]{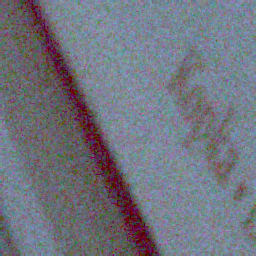}}& 
					{\includegraphics[width=0.120\textwidth]{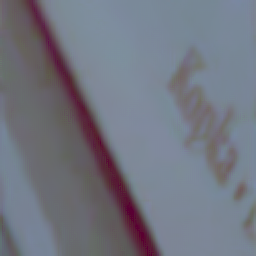}} & 
					{\includegraphics[width=0.120\textwidth]{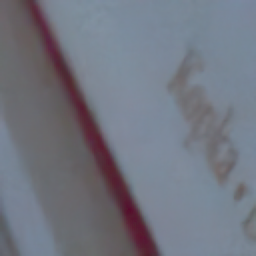}}&
					{\includegraphics[width=0.120\textwidth]{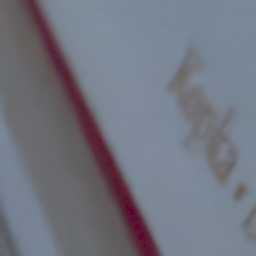}}&
					{\includegraphics[width=0.120\textwidth]{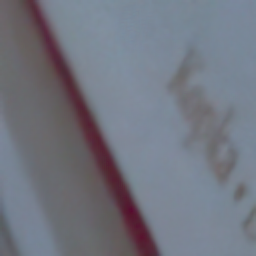}}&
					{\includegraphics[width=0.120\textwidth]{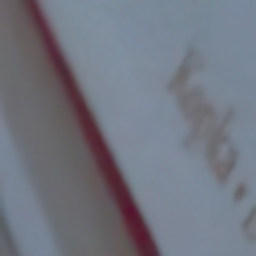}} &
					{\includegraphics[width=0.120\textwidth]{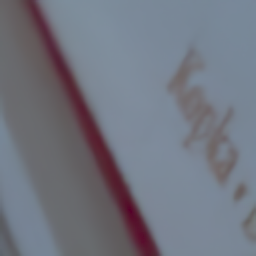}}&
					{\includegraphics[width=0.120\textwidth]{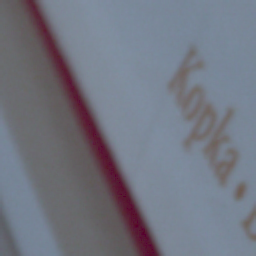}} \\
					29.43 / 0.8130 & 45.03 / 0.9945 & 45.03 / 0.9941 & \textbf{\color{blue}{46.09}} / \textbf{\color{blue}{0.9956}} & 45.62 / 0.9951 & 45.77 / 0.9952 & \textbf{\color{red}{46.86}} / \textbf{\color{red}{0.9967}} & PSNR / SSIM
				\end{tabular}%
			}
		\end{center}
		\caption{Blind raw image denoising results on images from the SIDD dataset. \textbf{(Best viewed with zoom-in)}}
		\label{fig:SIDD}
	\end{figure*}
	
	\begin{figure*}[t!]
		\small 
		\setlength\tabcolsep{1pt}
		\renewcommand\arraystretch{0.8}
		\centering
		\begin{center}
			{%
				\begin{tabular}{cccccccc}
					\makebox[0.120\textwidth][c]{Input} & 
					\makebox[0.120\textwidth][c]{VST+BM3D~\cite{BM3D}} &
					\makebox[0.120\textwidth][c]{NBR2NBR~\cite{CVPR21/NBR2NBR}} & 
					\makebox[0.120\textwidth][c]{FBI~\cite{CVPR21/FBI}} & 
					\makebox[0.120\textwidth][c]{B2U~\cite{CVPR22/B2U}}  & 
					\makebox[0.120\textwidth][c]{DCD~\cite{ICCV23/DCD}} & 
					\makebox[0.120\textwidth][c]{YOND} & 
					\makebox[0.120\textwidth][c]{Reference} \\
					{\includegraphics[width=0.120\textwidth,trim=128pt 256pt 128pt 0pt, clip]{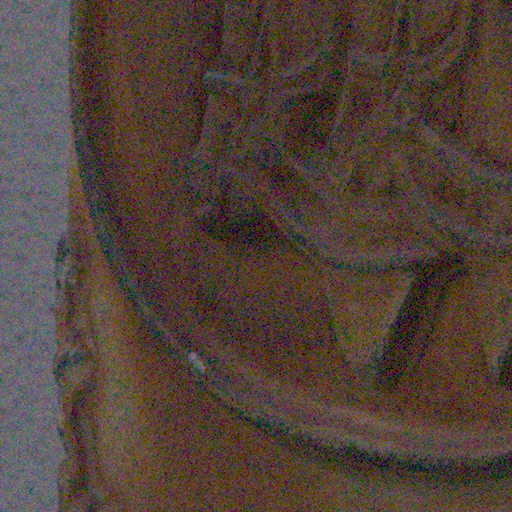}}& 
					{\includegraphics[width=0.120\textwidth,trim=128pt 256pt 128pt 0pt, clip]{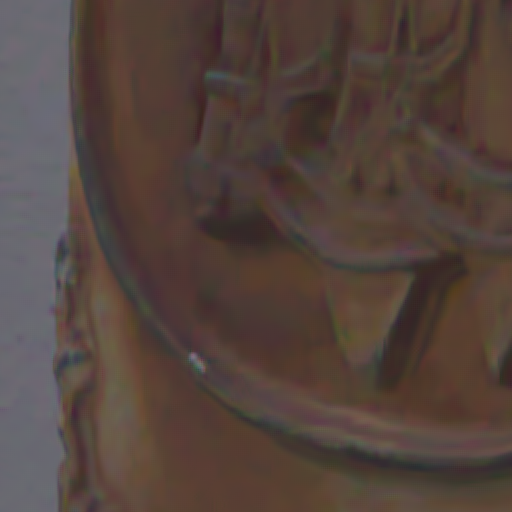}} & 
					{\includegraphics[width=0.120\textwidth,trim=128pt 256pt 128pt 0pt, clip]{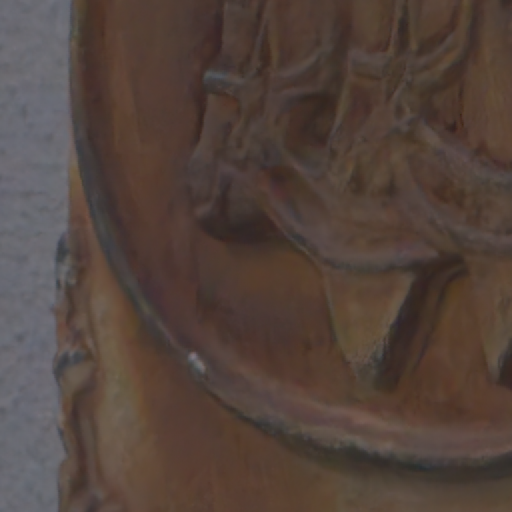}}&
					{\includegraphics[width=0.120\textwidth,trim=128pt 256pt 128pt 0pt, clip]{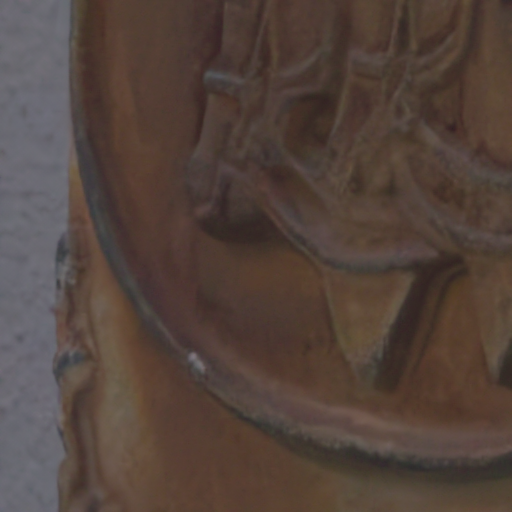}}&
					{\includegraphics[width=0.120\textwidth,trim=128pt 256pt 128pt 0pt, clip]{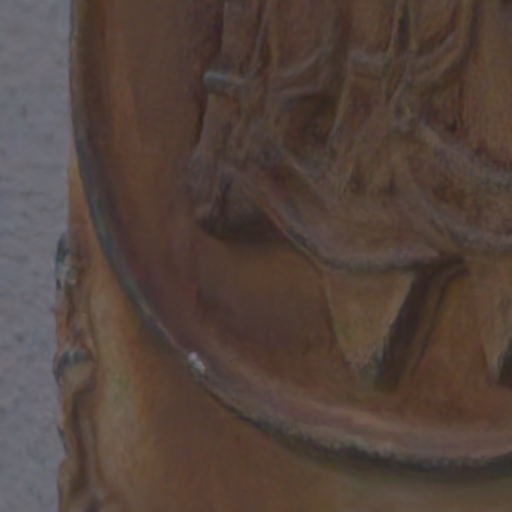}}&
					{\includegraphics[width=0.120\textwidth,trim=128pt 256pt 128pt 0pt, clip]{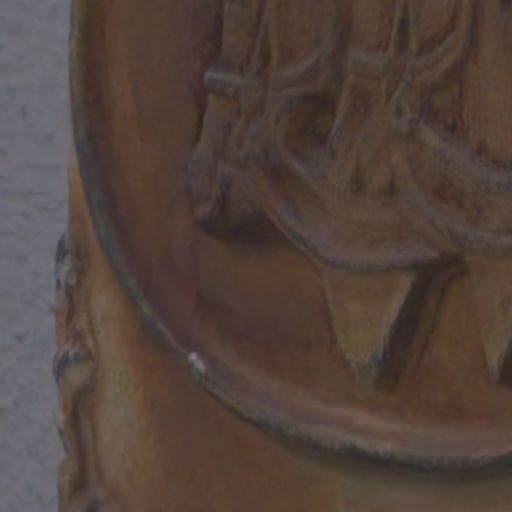}}&
					{\includegraphics[width=0.120\textwidth,trim=128pt 256pt 128pt 0pt, clip]{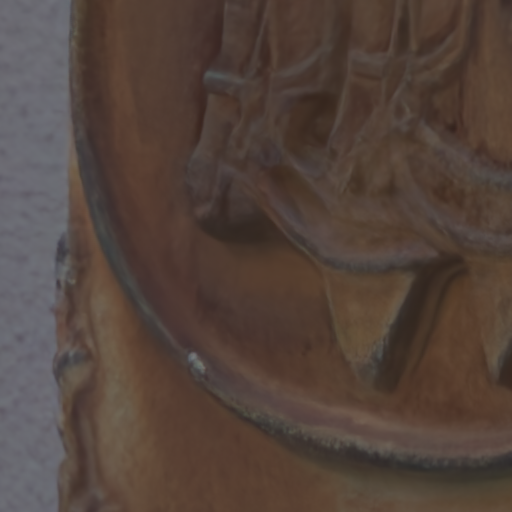}}&
					{\includegraphics[width=0.120\textwidth]{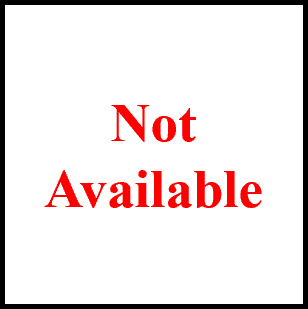}} \\
					{37.11} / {0.8838} & {46.92} / {0.9891} & {48.82} / {0.9924} & {48.77} / {0.9922} & {48.94} / \textbf{\color{blue}0.9926} & \textbf{\color{blue}48.98} / \textbf{\color{blue}0.9926} & \textbf{\color{red}49.24} / \textbf{\color{red}0.9931} & PSNR / SSIM \\
					\addlinespace[2pt]
					{\includegraphics[width=0.120\textwidth,trim=0pt 64pt 256pt 192pt, clip]{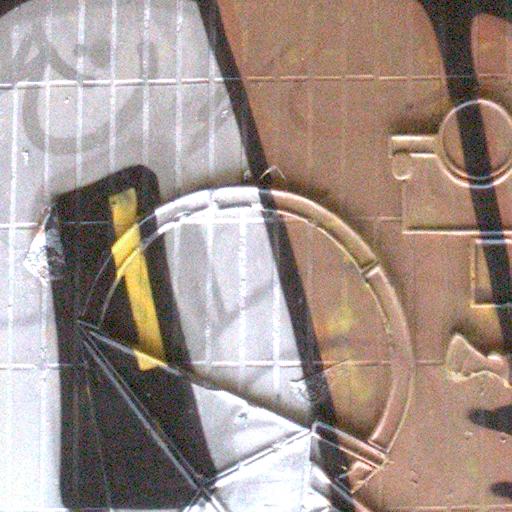}}& 
					{\includegraphics[width=0.120\textwidth,trim=0pt 64pt 256pt 192pt, clip]{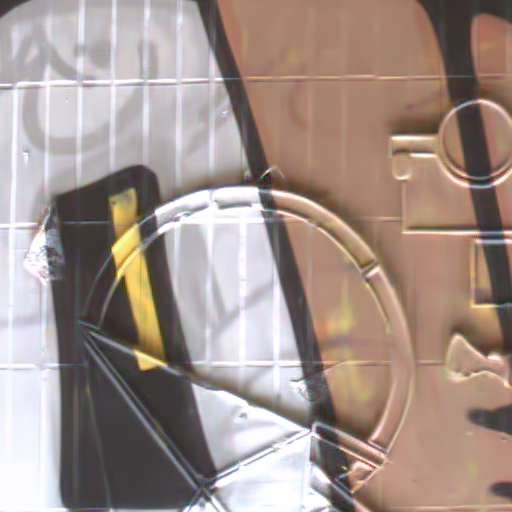}} & 
					{\includegraphics[width=0.120\textwidth,trim=0pt 64pt 256pt 192pt, clip]{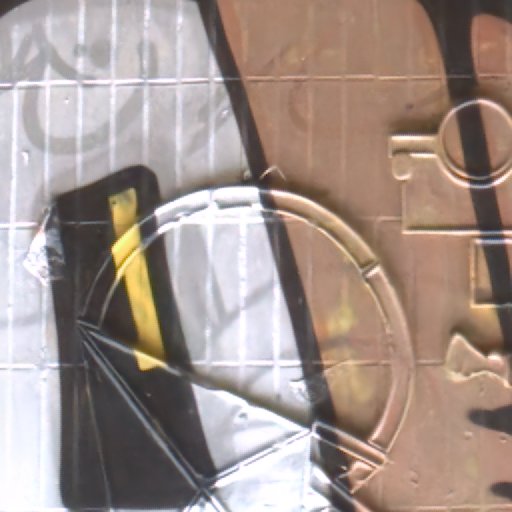}}&
					{\includegraphics[width=0.120\textwidth,trim=0pt 64pt 256pt 192pt, clip]{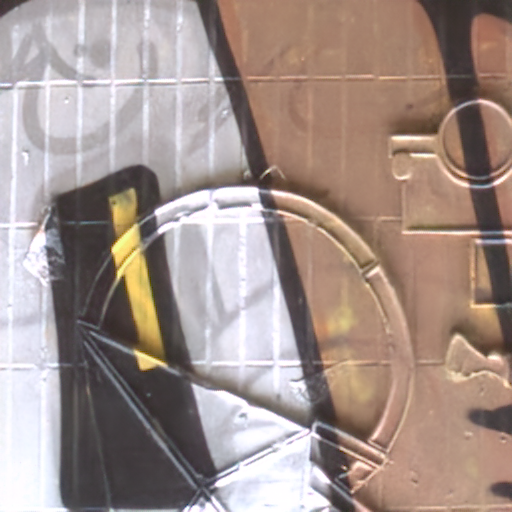}}&
					{\includegraphics[width=0.120\textwidth,trim=0pt 64pt 256pt 192pt, clip]{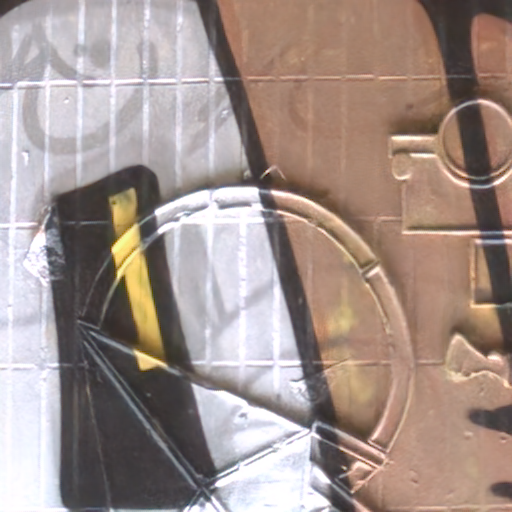}}&
					{\includegraphics[width=0.120\textwidth,trim=0pt 64pt 256pt 192pt, clip]{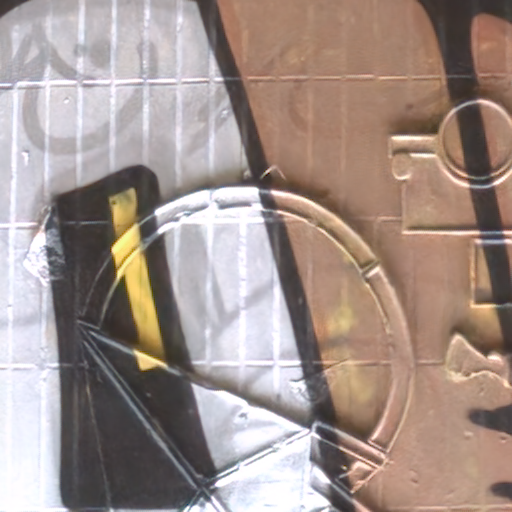}}&
					{\includegraphics[width=0.120\textwidth,trim=0pt 64pt 256pt 192pt, clip]{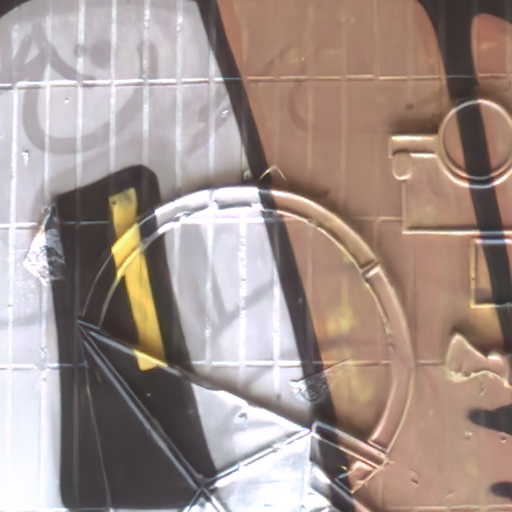}}&
					{\includegraphics[width=0.120\textwidth]{DND/GT}} \\
					{30.39} / {0.9372} & {34.13} / {0.9827} & {32.97} / {0.9815} & {33.41} / {0.9829} & \textbf{\color{blue}34.55} / \textbf{\color{blue}0.9854} & 34.10 / 0.9846 & \textbf{\color{red}35.28} / \textbf{\color{red}0.9873} & PSNR / SSIM \\
				\end{tabular}%
			}
		\end{center}
		\caption{Blind raw image denoising results on images from the DND dataset. \textbf{(Best viewed with zoom-in)}}
		\label{fig:DND}
	\end{figure*}
	
	\subsection{Comparison with Calibration-based Methods}
	Table~\ref{tab:nm} compares the denoising performance of calibration-based methods with YOND. We label the performance of P-G and ELD without camera-specific calibration as ``Blind" to facilitate a fair comparison of the blind denoising capability. The ``Blind" results are obtained by applying denoisers trained with noise parameters from another camera. Specifically, in the ELD dataset, we swap the noise parameters of SonyA7S2 and NikonD850, and in the LRID dataset, we replace the noise parameters of the IMX686 sensor with those of NikonD850.
	
	YOND achieves state-of-the-art scores in PSNR and competitive scores in SSIM. Figure~\ref{fig:ELD-compare} and Figure~\ref{fig:IMX686-compare} present some representative denoising results on the ELD and LRID datasets. Calibration-based methods with correct noise parameters generally outperform their blind counterparts. Some anomalies are observed at high ISOs in the ELD dataset, indicating potential inconsistencies between calibrated and actual noise parameters.
	P-G often leaves residual noise and artifacts, with the blind version particularly prone to under-denoising or excessive blurring due to noise parameter mismatches. 
	ELD accounts for black level error caused by dark current noise, however, mismatches with real-world noise parameters frequently lead to color bias, especially in the blind version.
	In contrast, YOND demonstrates superior denoising performance, with exact colors and clear details. Notably, YOND (UNet) shares the same network structure as calibration-based methods. Despite lacking SNR priors as precise guidance, the results of YOND (UNet) are on par with or even better than those of calibration-based methods, demonstrating the superiority of the YOND.
	
	Our experiments suggest that both camera settings and physical environments can interfere with imaging systems, thereby compromising the effectiveness of calibration-based methods. Developing blind raw denoising methods to flexibly adapt the changing environments is practical and necessary. YOND has surpassed calibration-based methods in most scenarios without any camera-specific data, demonstrating the practicality of our method.
	
	\begin{table*}[t]
		\footnotesize
		\setlength\tabcolsep{4pt}
		\caption{The blind denoising comparison of different self-supervised methods on the LRID dataset~\cite{TPAMI23/PMN}}\label{tab:calib}
		\centering
			\begin{tabular}{l|c|cc|cc|cc|cc}
				\Xhline{0.8pt}\rule{0pt}{8pt}\multirow{2.3}{*}{\makecell[c]{Method}} &
				\multirow{2.3}{*}{\makecell[c]{Blind}} & \multicolumn{2}{c}{Calibrated P-G Noise~\cite{P-G}} \vline & \multicolumn{2}{c}{Calibrated ELD Noise~\cite{TPAMI21/ELD}} \vline& \multicolumn{2}{c}{Calibrated SFRN Noise~\cite{ICCV21/SFRN}} \vline& \multicolumn{2}{c}{Real Noise~\cite{TPAMI23/PMN}}\\
				\Xcline{3-10}{0.4pt}
				\rule{0pt}{8pt} 
				& & ISO-6400 & ISO-12800 & ISO-6400 & ISO-12800 & ISO-6400 & ISO-12800 & ISO-6400 & ISO-12800\\
				\Xhline{0.6pt}\rule{0pt}{9pt}N2C & \XSolidBrush & {47.70} / \textbf{\color{blue}0.989} & \textbf{\color{blue}46.21} / \textbf{\color{red}0.984} & {47.68} / \textbf{\color{blue}0.989} & \textbf{\color{blue}46.18} / \textbf{\color{blue}0.984} & {47.46} / \textbf{\color{blue}0.989} & \textbf{\color{blue}46.00} / \textbf{\color{red}0.985} & \textbf{\color{blue}47.77} / \textbf{\color{red}0.990} & \textbf{\color{red}46.11} / \textbf{\color{red}0.985}\\
				\Xhline{0.4pt}\rule{0pt}{9pt}VST+BM3D~\cite{BM3D} & \CheckmarkBold & {46.49} / {0.984} & {44.09} / {0.970} & {46.48} / {0.984} & {44.10} / {0.970} & {46.19} / {0.981} & {43.55} / {0.964} & {45.63} / {0.981} & {43.16} / {0.963}\\
				N2N~\cite{ICML18/N2N} & \CheckmarkBold & {47.47} / {0.987} & {45.16} / {0.971} & {47.47} / {0.987} & {45.15} / {0.971} & {47.34} / {0.986} & {44.90} / {0.969} & {46.74} / {0.986} & {44.43} / {0.966}\\
				N2V~\cite{CVPR19/N2V} & \CheckmarkBold & {46.48} / {0.984} & {44.23} / {0.962} & {46.48} / {0.984} & {44.23} / {0.962} & {46.25} / {0.982} & {43.76} / {0.956} & {45.70} / {0.982} & {43.20} / {0.954}\\
				NBR2NBR~\cite{CVPR21/NBR2NBR} & \CheckmarkBold & {47.20} / {0.987} & {45.01} / {0.969} & {47.20} / {0.987} & {45.00} / {0.969} & {46.99} / {0.986} & {44.65} / {0.965} & {45.78} / {0.969} & {41.73} / {0.904}\\
				FBI~\cite{CVPR21/FBI} & \CheckmarkBold & {46.77} / {0.984} & {43.68} / {0.961} & {46.78} / {0.984} & {43.70} / {0.961} & {46.55} / {0.983} & {43.43} / {0.959} & {45.70} / {0.966} & {41.81} / {0.904}\\
				B2U~\cite{CVPR22/B2U} & \CheckmarkBold & {47.75} / {0.988} & {45.54} / {0.976} & {47.75} / {0.988} & {45.53} / {0.976} & \textbf{\color{blue}47.51} / {0.987} & {45.09} / {0.971} & {46.78} / {0.986} & {44.53} / {0.968}\\
				DCD~\cite{ICCV23/DCD} & \CheckmarkBold & \textbf{\color{blue}47.76} / \textbf{\color{blue}0.989} & {45.73} / {0.980} & \textbf{\color{blue}47.76} / \textbf{\color{blue}0.989} & {45.73} / {0.980} & \textbf{\color{blue}47.51} / {0.987} & {45.26} / {0.976} & {46.76} / {0.986} & {44.64} / {0.972}\\
				YOND & \CheckmarkBold & \textbf{\color{red}48.94} / \textbf{\color{red}0.991} & \textbf{\color{red}47.15} / \textbf{\color{red}0.984} & \textbf{\color{red}48.95} / \textbf{\color{red}0.991} & \textbf{\color{red}47.14} / \textbf{\color{red}0.984} & \textbf{\color{red}48.62} / \textbf{\color{red}0.990} & \textbf{\color{red}46.43} / \textbf{\color{blue}0.981} & \textbf{\color{red}47.82} / \textbf{\color{blue}0.989} & \textbf{\color{blue}45.67} / \textbf{\color{blue}0.977}\\
				\Xhline{0.8pt}
			\end{tabular}
			\begin{flushleft}
				$^1$ The \textbf{\color{red}red} color indicates the best results and the \textbf{\color{blue}blue} color indicates the second-best results. \\
				$^2$ ``Blind" represents ``no camera-specific training" here. N2C represents the supervised denoising results trained on corresponding datasets.
			\end{flushleft}
		\vspace{-6pt}
	\end{table*}
	
	\begin{table}[t]
		\footnotesize
		\setlength\tabcolsep{8pt}
		\caption{The comparison of different self-supervised methods on SIDD dataset~\cite{CVPR18/SIDD} and DND dataset~\cite{CVPR17/DND}}\label{tab:selfsup}
		\centering
			\begin{tabular}{l|cc|cc}
				\Xhline{0.8pt}\rule{0pt}{8pt}
				& \multicolumn{2}{c}{SIDD Dataset} \vline & \multicolumn{2}{c}{DND Dataset} \\
				\Xcline{2-5}{0.4pt}\rule{0pt}{8pt}
				\multirow{-2.3}{*}{Methods} & 
				\makebox[0.160\linewidth][c]{PSNR / SSIM} & \makebox[0.040\linewidth][c]{Blind} &
				\makebox[0.160\linewidth][c]{PSNR / SSIM} & \makebox[0.040\linewidth][c]{Blind} \\
				\Xhline{0.6pt}\rule{0pt}{9pt}VST+BM3D~\cite{BM3D} & 49.79 / 0.987 & \Checkmark & 46.52 / 0.965 & \Checkmark\\
				N2N~\cite{ICML18/N2N} & 51.29 / 0.991 & \XSolidBrush & 48.13 / 0.979 & \Checkmark\\
				N2V~\cite{CVPR19/N2V} & 50.46 / 0.990 & \XSolidBrush & 47.29 / 0.977 & \Checkmark\\
				NBR2NBR~\cite{CVPR21/NBR2NBR} & 51.06 / 0.991 & \XSolidBrush &  47.75 / 0.978 & \Checkmark\\
				FBI~\cite{CVPR21/FBI} & 51.14 / 0.991 & \XSolidBrush & 48.09 / 0.979 & \Checkmark\\
				B2U~\cite{CVPR22/B2U} & 51.36 / \textbf{0.992} & \XSolidBrush & 48.19 / 0.980 & \Checkmark\\
				DCD~\cite{ICCV23/DCD} & 51.40 / \textbf{0.992} & \XSolidBrush & 48.24 / 0.980 & \Checkmark\\
				YOND & \textbf{51.60} / 0.990 & \Checkmark & \textbf{48.57} / \textbf{0.981} & \Checkmark\\
				\Xhline{0.8pt}
			\end{tabular}
			\begin{flushleft}
				$^1$ The \textbf{bold} scores denote the best results. \\
				$^2$ ``Blind" represents ``no camera-specific training" here. \\
			\end{flushleft}
			\vspace{-6pt}
	\end{table}
	
	\subsection{Comparison with Self-supervised Methods}
	
	Table~\ref{tab:calib} shows the blind denoising performance of self-supervised methods on the LRID dataset without camera-specific training. We additionally synthesize three datasets on LRID based on various calibrated noise models for comparison. N2C, \ie, Noise-to-Clean, represents the supervised denoising results, requiring camera-specific training on the corresponding datasets. For example, N2C on the calibrated P-G synthetic dataset is trained with P-G noise, while N2C on the Real Noise dataset is trained with real noise (paired real data). For the convenience of presentation, we have combined the results for indoor and outdoor scenes at the same ISO. Using N2C as a baseline, we find that most self-supervised methods exhibit a performance gap of over 1 dB compared to supervised denoising without camera-specific training. Some advanced self-supervised methods, \ie, B2U and DCD, achieve results close to supervised denoising on synthetic datasets. However, as the noise model approaches real-world conditions, the blind denoising performance of self-supervised methods significantly degrades. In contrast, YOND demonstrates state-of-the-art blind denoising capabilities, achieving competitive results even on real noise.
	
	Table~\ref{tab:selfsup} shows the blind denoising performance of self-supervised methods on the SIDD and DND datasets, where most of them have undergone camera-specific training on SIDD. We present some representative denoising results in Figure~\ref{fig:SIDD} and Figure~\ref{fig:DND}. 
	Traditional VST+BM3D tends to produce excessively blurry denoised images. N2V performs poorly due to the loss of central pixel information. N2N lacks practicality due to its reliance on a large number of paired noisy images. NBR2NBR usually exhibits artifacts in high-frequency regions. FBI leaves residual noise due to inaccurate noise estimation. B2U and DCD struggle to recover sharp details under high noise level. YOND demonstrates outstanding visual results in most scenes, providing clean images with clear details. 
	
	It is important to emphasize that collecting large, diverse noisy data and conducting camera-specific denoiser training is impractical for individual users. In contrast, YOND has no camera-specific data dependency, only requiring training once on camera-independent synthetic data. 
	YOND achieves pleasing denoising performance on 9 different cameras from the SIDD datasets and DND datasets, demonstrating the superior practicality of our method.
	
	\begin{table*}[t]
		\centering
		\scriptsize
		\setlength\tabcolsep{2.7pt}
		\caption{The comparison of different noise estimation methods. The noisy images are synthesized by the calibrated noise parameters of a smartphone and a DSLR camera respectively\label{tab:NoiseEstimation}}
		\begin{tabular}{c|c|cc|ll|ll|ll|ll|ll}
			\Xhline{0.8pt}\rule{0pt}{8pt}
			& & 
			\multicolumn{2}{c}{\makebox[0.100\linewidth][c]{Noise Parameters}} \vline& \multicolumn{2}{c}{Foi~\cite{P-G}} \vline& \multicolumn{2}{c}{PGE~\cite{CVPR21/FBI}} \vline& \multicolumn{2}{c}{Zou~\cite{CVPR22/FG-ELD}} \vline& 
			\multicolumn{2}{c}{CNE (Coarse)} \vline&
			\multicolumn{2}{c}{CNE (Fine)} \\
			\Xcline{3-14}{0.4pt}\rule{0pt}{8pt}
			\multirow{-2.5}{*}{Camera} & \multirow{-2.5}{*}{\makecell{ISO\\($\times$ratio)}} & 
			\makecell[c]{$\alpha$} & \makecell[c]{$\sigma$} & \makecell[c]{$\alpha$} & \makecell[c]{$\sigma$} & \makecell[c]{$\alpha$} & \makecell[c]{$\sigma$} & \makecell[c]{$\alpha$} & \makecell[c]{$\sigma$} &
			\makecell[c]{$\alpha$} & \makecell[c]{$\sigma$} & \makecell[c]{$\alpha$} & \makecell[c]{$\sigma$} \\
			
			\Xhline{0.6pt}\rule{0pt}{8pt}
			& 800 & \cellcolor{lightgray}{1.10e-3} & \cellcolor{lightgray}{2.20e-3} & 1.21e-3$_{\,11\%}$ & 1.61e-3$_{\,27\%}$ & \textbf{\color{blue}1.13e-3$_\mathbf{\,3\%}$} & 2.43e-3$_{\,11\%}$ & 9.29e-4$_{\,16\%}$ & \textbf{\color{blue}2.25e-3$_\mathbf{\,3\%}$} & \textbf{\color{blue}1.13e-3$_\mathbf{\,3\%}$} & 2.82e-3$_{\,29\%}$ & \textbf{\color{red}1.12e-3$_\mathbf{\,2\%}$} & \textbf{\color{red}2.16e-3$_\mathbf{\,2\%}$}\\
			& 1600 & \cellcolor{lightgray}{2.30e-3} & \cellcolor{lightgray}{4.00e-3} & 2.60e-3$_{\,14\%}$ & 3.16e-3$_{\,21\%}$ & 2.52e-3$_{\,10\%}$ & \textbf{\color{blue}3.53e-3$_\mathbf{\,12\%}$} & 1.50e-3$_{\,35\%}$ & 4.50e-3$_{\,13\%}$ & \textbf{\color{blue}2.35e-3$_\mathbf{\,3\%}$} & \textbf{\color{blue}4.47e-3$_\mathbf{\,12\%}$} & \textbf{\color{red}2.32e-3$_\mathbf{\,1\%}$} & \textbf{\color{red}3.97e-3$_\mathbf{\,1\%}$}\\
			& 3200 & \cellcolor{lightgray}{4.60e-3} & \cellcolor{lightgray}{7.20e-3} & 4.97e-3$_{\,9\%}$ & 6.22e-3$_{\,14\%}$ & 5.52e-3$_{\,21\%}$ & 4.96e-3$_{\,32\%}$ & 2.52e-3$_{\,46\%}$ & \textbf{\color{red}7.20e-3$_\mathbf{\,0\%}$} & \textbf{\color{blue}4.65e-3$_\mathbf{\,2\%}$} & 7.67e-3$_{\,7\%}$ & \textbf{\color{red}4.62e-3$_\mathbf{\,1\%}$} & \textbf{\color{blue}7.17e-3$_\mathbf{\,1\%}$}\\
			\multirow{-4}{*}{Phone} & 6400 & \cellcolor{lightgray}{9.10e-3} & \cellcolor{lightgray}{1.30e-2} & 9.94e-3$_{\,10\%}$ & 8.70e-3$_{\,34\%}$ & 1.21e-2$_{\,34\%}$ & 5.93e-3$_{\,55\%}$ & 4.33e-3$_{\,53\%}$ & \textbf{\color{blue}1.27e-2$_\mathbf{\,3\%}$} & \textbf{\color{blue}9.15e-3$_\mathbf{\,1\%}$} & \textbf{\color{blue}1.33e-2$_\mathbf{\,3\%}$} & \textbf{\color{red}9.07e-3$_\mathbf{\,1\%}$} & \textbf{\color{red}1.30e-2$_\mathbf{\,0\%}$}\\
			\Xhline{0.6pt}\rule{0pt}{8pt}
			& 3200 & \cellcolor{lightgray}{1.90e-3} & \cellcolor{lightgray}{2.50e-3} & 2.07e-3$_{\,9\%}$ & \textbf{\color{blue}2.18e-3$_\mathbf{\,13\%}$} & 1.85e-3$_{\,3\%}$ & 3.05e-3$_{\,22\%}$ & 1.25e-3$_{\,35\%}$ & 2.84e-3$_{\,14\%}$ & \textbf{\color{blue}1.92e-3$_\mathbf{\,2\%}$} & 3.09e-3$_{\,24\%}$ & \textbf{\color{red}1.91e-3$_\mathbf{\,1\%}$} & \textbf{\color{red}2.46e-3$_\mathbf{\,2\%}$}\\
			& 6400 & \cellcolor{lightgray}{3.85e-3} & \cellcolor{lightgray}{4.50e-3} & 4.28e-3$_\mathbf{\,12\%}$ & 4.09e-3$_{\,10\%}$ & 3.95e-3$_{\,3\%}$ & \textbf{\color{blue}4.26e-3$_\mathbf{\,6\%}$} & 2.51e-3$_{\,35\%}$ & 4.79e-3$_{\,7\%}$ & \textbf{\color{blue}3.89e-3$_\mathbf{\,2\%}$} & 5.14e-3$_{\,15\%}$ & \textbf{\color{red}3.86e-3$_\mathbf{\,1\%}$} & \textbf{\color{red}4.45e-3$_\mathbf{\,2\%}$}\\
			& 12800 & \cellcolor{lightgray}{7.70e-3} & \cellcolor{lightgray}{9.00e-3} & 8.65e-3$_{\,13\%}$ & \textbf{\color{blue}8.70e-3$_\mathbf{\,4\%}$} & 8.85e-3$_{\,15\%}$ & 5.66e-3$_{\,38\%}$ & 4.17e-3$_{\,46\%}$ & 7.89e-3$_{\,13\%}$ & \textbf{\color{blue}7.74e-3$_\mathbf{\,1\%}$} & 9.45e-3$_{\,6\%}$ & \textbf{\color{red}7.70e-3$_\mathbf{\,0\%}$} & \textbf{\color{red}8.96e-3$_\mathbf{\,1\%}$}\\
			\multirow{-4}{*}{DSLR} & 25600 & \cellcolor{lightgray}{1.55e-2} & \cellcolor{lightgray}{1.63e-2} & 1.77e-2$_{\,15\%}$ & 1.72e-2$_{\,6\%}$ & 1.87e-2$_{\,21\%}$ & 5.89e-3$_{\,64\%}$ & 6.02e-3$_{\,62\%}$ & 1.72e-2$_{\,6\%}$ & \textbf{\color{blue}1.56e-2$_\mathbf{\,1\%}$} & \textbf{\color{blue}1.67e-2$_\mathbf{\,3\%}$} & \textbf{\color{red}1.54e-2$_\mathbf{\,1\%}$} & \textbf{\color{red}1.63e-2$_\mathbf{\,0\%}$}\\
			\Xhline{0.8pt}
		\end{tabular}
		
	\footnotesize
	\begin{flushleft}
		$^1$ The \sethlcolor{lightgray}\hl{gray values} denote the true noise parameters, the \textbf{\color{red}red} color indicates the best noise estimation results and the \textbf{\color{blue}blue} color indicates the second-best noise estimation results.\\
		$^2$ $\alpha$ and $\sigma$ are mean values of estimated noise parameters, where we indicate the deviation (\%) in the subscripts. ``ISO ($\times$ratio)" denotes the approximate camera ISO (including digital gain) corresponding to the normalized noise parameters.
	\end{flushleft}
	\vspace{-6pt}
\end{table*}

\subsection{Ablation Study}
\subsubsection{\textbf{Ablation on CNE}}

\begin{figure}[t]
	\centering
	\includegraphics[width=\linewidth,trim=5 0 0 0,clip]{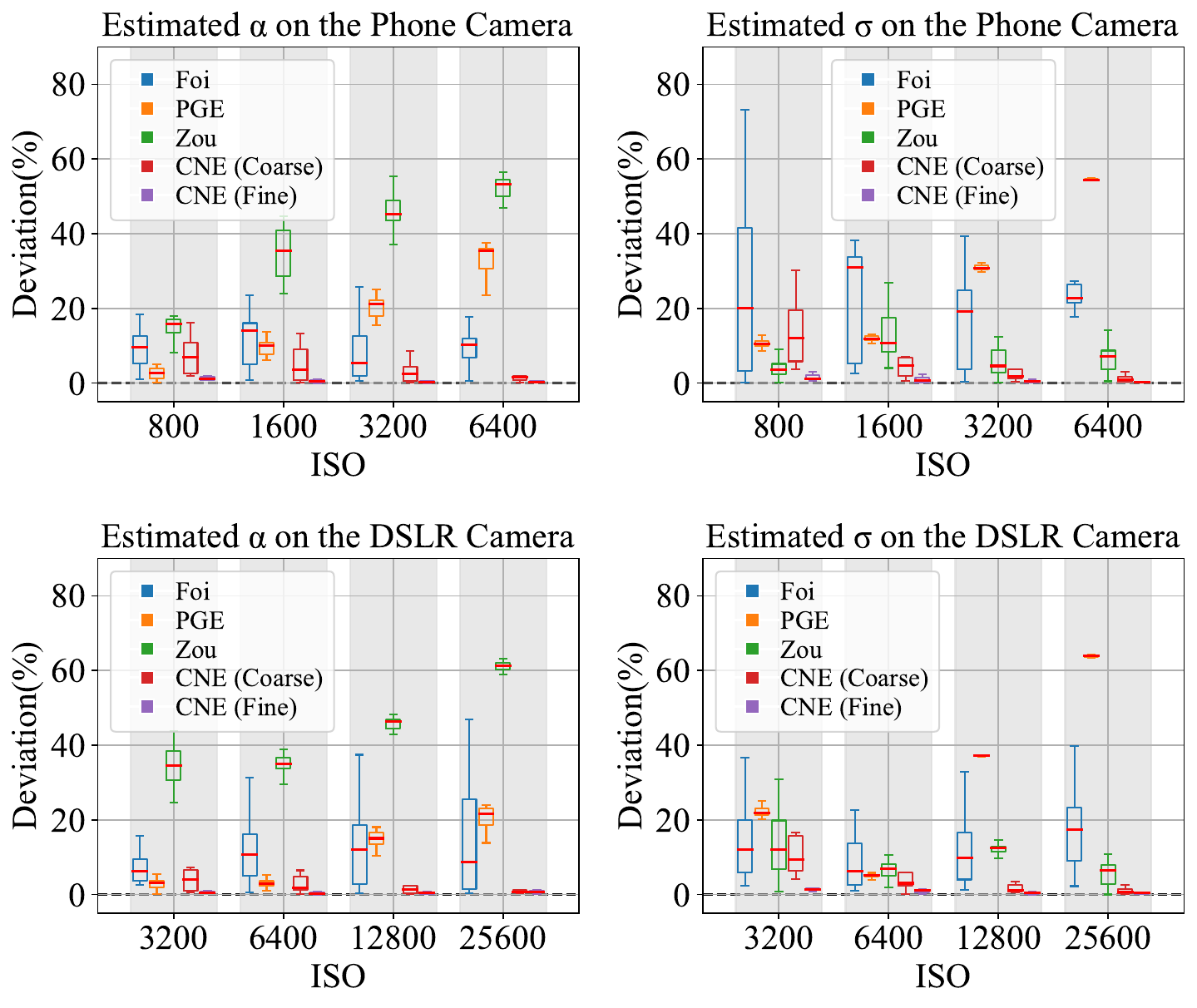}
	\caption{Error analysis of noise estimation results at various ISO of smartphone camera and DSLR camera. The deviation (y-axis) represents the ratio of noise estimation results to the true noise parameters. The black dashed line corresponding to ``0" indicates the estimation without deviation, and the gap from the black dashed line represents the deviation. The red solid line within each box represents the median of estimation results. The length of the box represents the stability of the estimation.}
	\label{fig:NoiseEstimation}
	\vspace{-6pt}
\end{figure}
%

For noise estimation, we calibrate noise parameters using results from a smartphone camera (with IMX686 sensor) and a DSLR camera (SonyA7S2). To avoid interference from residual noise in the dataset, we select reference images from the LRID dataset to synthesize noisy images. Poisson-Gaussian model is used for generating noise based on the noise parameters at various ISOs. We compare our approach with three representative noise estimation methods: Foi~\cite{P-G}, PGE~\cite{CVPR21/FBI}, and Zou~\cite{CVPR22/FG-ELD}. Foi is a robust traditional method for single-image noise estimation, which is still highly competitive today. PGE is a learning-based method used to provide accurate noise parameters for the VST in FBI-Net~\cite{CVPR21/FBI}. Zou is a camera noise calibration method based on contrastive learning. In the experiment, the estimated $\sigma^2$ of some methods may be negative sometimes. In this case, we will set $\sigma=0$ to avoid obvious errors.

In Table~\ref{tab:NoiseEstimation} and Figure~\ref{fig:NoiseEstimation}, we visually present the accuracy of different noise estimation methods using box plots~\cite{boxplot} and mean error, respectively. The box plots focus on the stability of single-image noise parameter estimation, while the mean error table highlights the accuracy of camera noise parameter estimation. Most methods show similar tendencies in both statistics, with Zou performing slightly better in mean error. 
Foi is greatly degraded in complex scenarios with rich textures, resulting in estimated results inaccurate and unstable. 
Both PGE and Zou are learning-based methods, thus they both have serious data dependency. PGE excels at estimating \(\alpha\), while Zou is good at estimating \(\sigma\), however, both of them exhibit notable bias in the parameters they are less proficient with. Unlike Foi, which shows consistent performance across different ISOs, PGE and Zou perform significantly better at lower ISOs compared to higher ISOs. 
In contrast, the coarse estimation in CNE already shows some advantages. The full CNE exhibits both exceptional accuracy and remarkable stability across different ISO levels, demonstrating its superiority in noise estimation.

\begin{table}[t!]
	\footnotesize
	\setlength\tabcolsep{4pt}
	\caption{Ablation studies of CNE on SIDD dataset}\label{tab:CNE}
	\centering
		\begin{tabular}{lc|lc}
			\Xhline{0.8pt}
			\multicolumn{2}{c}{Single-Step Estimation}\vline & \multicolumn{2}{c}{Coarse-to-fine Estimation} \\
			\Xhline{0.4pt}\rule{0pt}{8pt}Method & \multicolumn{1}{c}{PSNR / SSIM}\vline & Method & PSNR / SSIM \\ 
			\Xhline{0.6pt}\rule{0pt}{8pt}CameraNLF~\cite{ICCV21/SFRN} & 49.29 / 0.9660 & CameraNLF + CNE & 51.42 / 0.9866 \\
			Foi~\cite{P-G} & 50.80 / 0.9875 &  Foi + CNE & 51.59 / \textbf{0.9895} \\
			PGE~\cite{CVPR21/FBI} & 50.24 / 0.9804 & PGE + CNE & 51.59 / 
			0.9888 \\
			Zou~\cite{CVPR22/FG-ELD} & 46.83 / 0.9622 & Zou + CNE & 50.84 / 
			0.9873 \\
			CNE (Coarse) & 50.95 / 0.9838 & CNE (Fine) & \textbf{51.60} / 
			\textbf{0.9895} \\
			\Xhline{0.8pt}
		\end{tabular}
	\begin{flushleft}
		$^1$ The \textbf{bold} scores denotes the best results.
	\end{flushleft}
\end{table}

\begin{figure}[t!]
	\footnotesize
	\setlength\tabcolsep{1pt}
	\centering
	\begin{tabular}{cccc}
		CameraNLF~\cite{ICCV21/SFRN} & Foi~\cite{P-G} & PGE~\cite{CVPR21/FBI} & Zou~\cite{CVPR22/FG-ELD} \\
		{\includegraphics[width=0.24\linewidth]{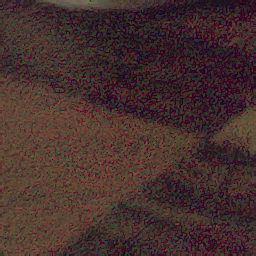}} &
		{\includegraphics[width=0.24\linewidth]{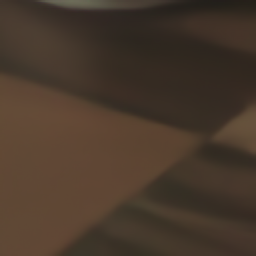}} &
		{\includegraphics[width=0.24\linewidth]{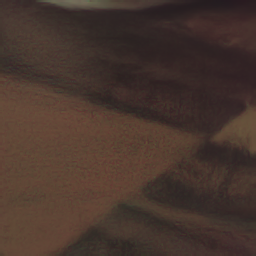}} & 
		{\includegraphics[width=0.24\linewidth]{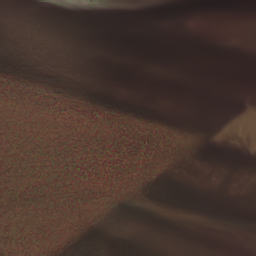}}  \\
		38.02 / 0.8763 & 45.75 / 0.9765 & 45.19 / 0.9747 & 44.74 / 0.9743 \\
		\addlinespace[2pt]
		CNE (Coarse) & CNE (Fine) & Noisy Image & Reference \\
		{\includegraphics[width=0.24\linewidth]{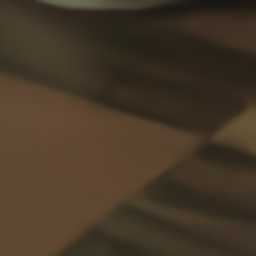}} &
		{\includegraphics[width=0.24\linewidth]{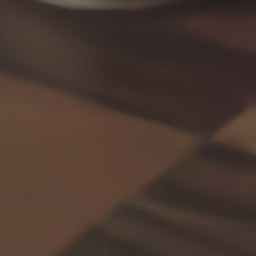}} &
		{\includegraphics[width=0.24\linewidth]{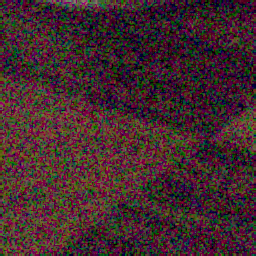}} & 
		{\includegraphics[width=0.24\linewidth]{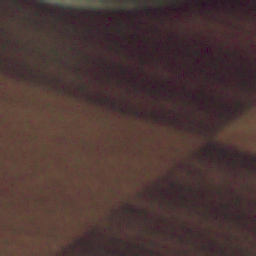}}  \\
		42.85 / 0.9446 & \textbf{47.72} / \textbf{0.9868} & 30.77 / 0.5841 & PSNR / SSIM \\
	\end{tabular}
	\caption{Visual comparison of the denoising results under different noise estimation methods. \textbf{(Best viewed with zoom-in)}}
	\label{fig:CNE}
\end{figure}

We further investigate the impact of noise estimation on the denoising performance of YOND in Table~\ref{tab:CNE} and Figure~\ref{fig:CNE}. To validate the robustness of the coarse-to-fine strategy, the denoising process is divided into two stages. In the coarse stage, we replace the coarse estimation results with noise parameters calculated by different estimation methods, followed by the fine estimation strategy to enhance accuracy. The noise calibration results are also included in the experiment, labeled as CameraNLF. Since SFRN~\cite{ICCV21/SFRN} reports inaccuracies in the noise parameters provided by the SIDD dataset, we adopt the recalibrated results from SFRN as CameraNLF.
As shown in Table~\ref{tab:CNE}, all noise parameter estimation methods significantly benefit from the coarse-to-fine strategy. Our CNE achieves the best results in both the coarse and fine stages, while other methods show similar performance after refinement, highlighting the robustness of CNE. Notably, the performance of CameraNLF in the coarse stage is expected to represent the upper bound on the SIDD dataset. However, the noise parameters provided by CameraNLF lead to poor denoising results. This anomaly underscores the complexity of noise calibration and further emphasizes the importance of YOND as a blind denoising method.



\subsubsection{\textbf{Ablation on EM-VST}}\label{sec:ab_BPC}
\begin{figure}[t]
	\footnotesize
	\setlength\tabcolsep{1pt}
	\centering
	\begin{tabular}{cccc}
		GAT+IAT~\cite{GAT} & GAT+UIAT~\cite{TIP13/VST} & EM-VST (Ours) & Reference \\
		{\includegraphics[width=0.24\linewidth]{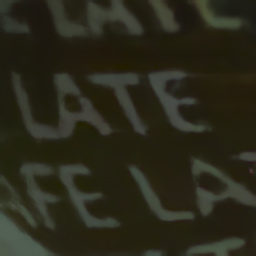}} &
		{\includegraphics[width=0.24\linewidth]{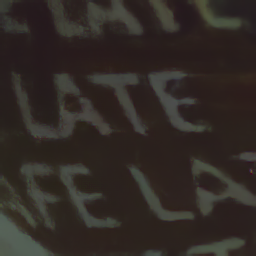}} &
		{\includegraphics[width=0.24\linewidth]{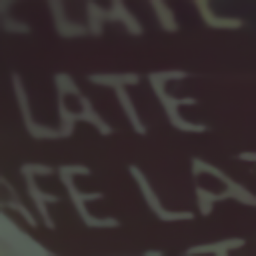}} & 
		{\includegraphics[width=0.24\linewidth]{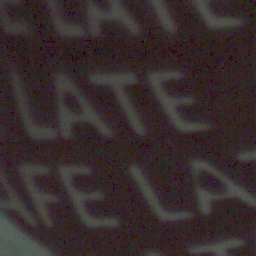}}  \\
		40.40 / 0.8568 & 42.88 / 0.9396 & \textbf{45.74} / \textbf{0.9761} & PSNR / SSIM\\
	\end{tabular}
	\caption{Visual comparison of the denoising results under different VST bias correction methods. \textbf{(Best viewed with zoom-in)}}
	\label{fig:ab_BPC}
\end{figure}

The ablation study on EM-VST is presented in Table~\ref{tab:BPC} and Figure~\ref{fig:ab_BPC}. We consider GAT+IAT~\cite{GAT}, GAT+UIAT~\cite{TIP13/VST}, and EM-VST for comparison. GAT+IAT itself does not involve bias correction, causing the denoised images to exhibit significant color bias in low-light regions due to the bias introduced by VST. GAT+UIAT corrects bias based on denoising results, relying on perfect denoising for unbiased correction, which obviously lacks accuracy. Benefiting from the EM-VST, our method accurately corrects the bias introduced by VST before denoising. EM-VST significantly promotes the color accuracy of denoised images in low-light regions.

\begin{table}[t]
	\footnotesize
	\caption{Ablation studies of EM-VST on SIDD dataset}\label{tab:BPC}
	\centering
		\begin{tabular}{lccc}
			\toprule
			& GAT+IAT~\cite{GAT} & GAT+UIAT~\cite{TIP13/VST} & EM-VST (Ours) \\
			\multirow{-2}{*}{Method} & PSNR / SSIM & PSNR / SSIM & PSNR / SSIM \\
			\midrule
			YOND & 49.27 / 0.9719 & 50.94 / 0.9849 & \textbf{51.60} / \textbf{0.9895} \\
			\bottomrule
		\end{tabular}
\end{table}

\subsubsection{\textbf{Ablation on SNR-Net}}\label{sec:ab_SNR-Net}
\begin{figure}[t!]
\footnotesize
\setlength\tabcolsep{1pt}
\centering
\begin{tabular}{cccc}
	UNet~\cite{Unet} & w/o SNR-guided & SNR-Net (Ours) & Reference \\
	{\includegraphics[width=0.24\linewidth]{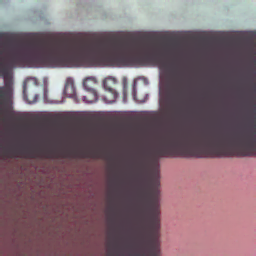}} &
	{\includegraphics[width=0.24\linewidth]{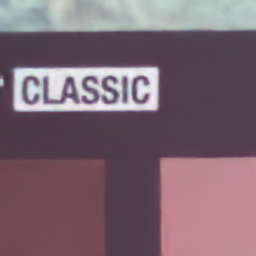}} &
	{\includegraphics[width=0.24\linewidth]{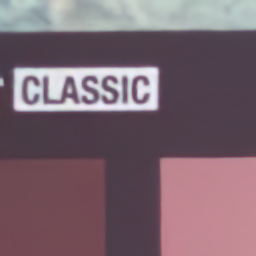}} & 
	{\includegraphics[width=0.24\linewidth]{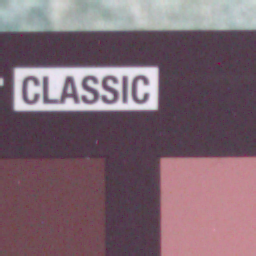}}  \\
	34.01 / 0.9681 & 35.36 / 0.9768 & \textbf{36.20} / \textbf{0.9772} & PSNR / SSIM\\
\end{tabular}
\caption{Visual comparison of the denoising results under different network structures. \textbf{(Best viewed with zoom-in)}}
\label{fig:ab_SNR}
\end{figure}

The ablation study on SNR-Net is presented in Table~\ref{tab:SNR} and Figure~\ref{fig:ab_SNR}. We consider three structures for comparison. The UNet structure, aligning with calibration-based methods, exhibits limited denoising performance due to its modest network capacity. 
The absence of the SNR-guided branch in SNR-Net does not diminish the network capacity but tends to result in noticeable artifacts. Benefiting from the SNR-guided branch, SNR-Net precisely removes noise and artifacts, showing the best visual quality.
We observe that SNR-Net performs less satisfactorily in terms of SSIM. The disadvantage comes from the sensitivity of SNR-Net to the accuracy of noise estimation, which is introduced by the SNR-guided branch. The challenging cases of CNE significantly diminish our quantitative results, especially in the detail-sensitive SSIM. A detailed analysis of challenging cases will be presented in the next section.

\section{Discussion}\label{sec:discuss}
\subsection{{Challenging Cases and Flexible Solutions}}\label{sec:failure}
\subsubsection{\textbf{Challenging Cases}}
\begin{figure}[t!]
\footnotesize
\setlength\tabcolsep{1pt}
\centering
\begin{tabular}{cccc}
	Noisy Image & Coarse Result & Fine Result & Reference \\
	{\includegraphics[width=0.24\linewidth]{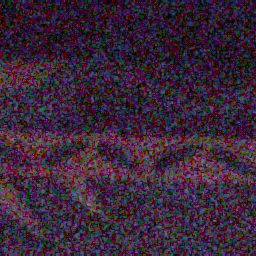}} &
	{\includegraphics[width=0.24\linewidth]{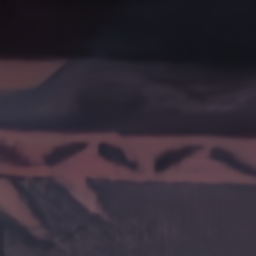}} &
	{\includegraphics[width=0.24\linewidth]{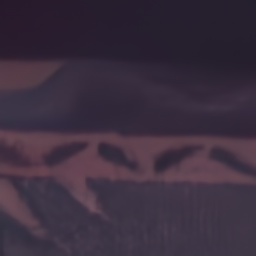}} & 
	{\includegraphics[width=0.24\linewidth]{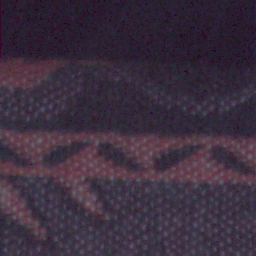}}  \\
	31.49 / 0.6085 & 40.73 / 0.8970 & \textbf{42.01} /\textbf{0.9517} & PSNR / SSIM \\
\end{tabular}
\caption{Visual results of a representative challenging case. \textbf{(Best viewed with zoom-in)}}
\label{fig:failure}
\end{figure}

We have conducted a comprehensive analysis for each denoised patch in the SIDD dataset. YOND achieves state-of-the-art results on 68.44\% patches, especially dominating in challenging scenes with strong noise. However, YOND sometimes encounters challenging cases in scenarios with rich textures, leading to a noticeable degradation in the detail-sensitive SSIM.
We have designed several practical corrections for scenarios with rich textures. CNE relies solely on the signals from flat regions for estimation and corrects noise estimation using denoised images. However, some extreme situations render practical corrections ineffective. In scenarios where flat regions are close to single-intensity patches, the errors of noise parameter fitting become substantial. When the entire image lacks flat regions, the flat region mask can only cover the regions with textures. The inclusion of textures within the flat region mask results in an overestimation in the coarse estimation, introducing a tendency towards blurring in the coarse denoised images. If the coarse denoised image is excessively blurry, the corrective capacity of the fine estimation becomes severely limited, resulting in blurry images as shown in Figure~\ref{fig:failure}. The cascade of failures constitutes the origins of challenging cases observed in YOND on the SIDD dataset.

\begin{table}[t!]
	\footnotesize
	\caption{Ablation studies of SNR-Net on SIDD Dataset}\label{tab:SNR}
	\centering
	\begin{tabular}{lccc}
		\toprule
		& UNet~\cite{Unet} & w/o SNR-guided & SNR-Net (Ours) \\
		\multirow{-2}{*}{Method} & PSNR / SSIM & PSNR / SSIM & PSNR / SSIM \\
		\midrule
		YOND & 50.66 / 0.9898 & 51.44 / \textbf{0.9911} & \textbf{51.60} / 0.9895 \\
		\bottomrule
	\end{tabular}
	\end{table}

The default settings enable YOND to outperform existing methods in most cases, while handling challenging cases is particularly essential in practice. A denoising method should be controllable to fulfill the slogan of ``You Only Need a Denoiser". Breaking data dependency has unlocked the powerful interactivity of YOND, allowing manual adjustments of various parameters to handle challenging cases. Next, we will we introduce two solutions to expand the applicability of YOND beyond its default settings.

\begin{figure}[t]
\footnotesize
\setlength\tabcolsep{1pt}
\centering
\subfloat[Flat Region Mask Adjustment]{\label{fig:discuss_mask}
	\includegraphics[width=\linewidth]{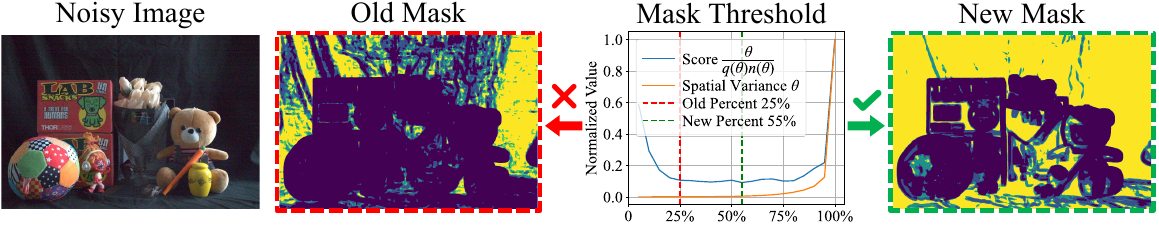}
}
\\
\vspace{-4pt}
\subfloat[Noise Parameter Adjustment]{\label{fig:discuss_CNE}
	\includegraphics[width=\linewidth]{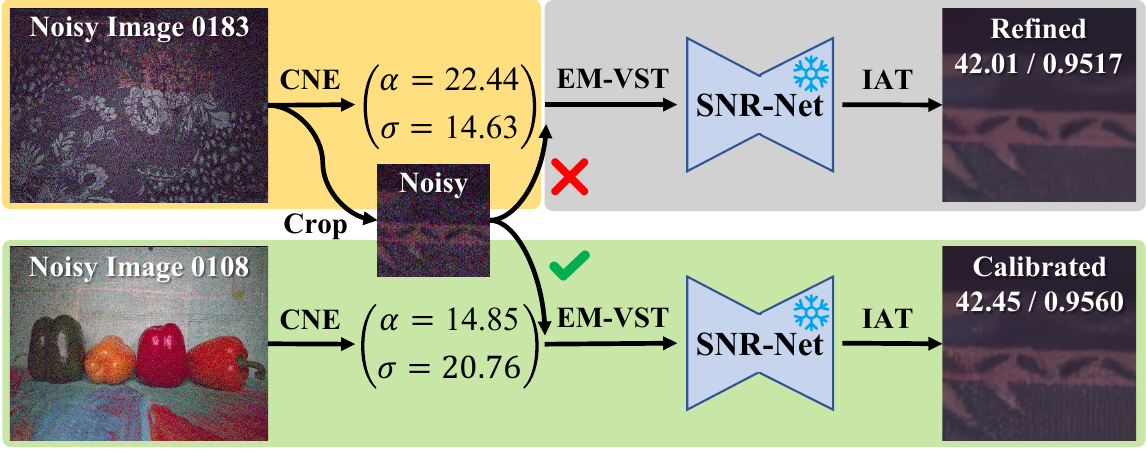}
}
\\
\vspace{-4pt}
\subfloat[Noise Level Adjustment]{\label{fig:discuss_snr}
	\begin{tabular}{ccccc}
		$\sigma_{\text{SNR}}\times$1.03 & $\sigma_{\text{SNR}}\times$0.98 & $\sigma_{\text{SNR}}\times$0.93 & $\sigma_{\text{SNR}}\times$0.88 & $\sigma_{\text{SNR}}\times$0.83\\
		{\includegraphics[width=0.19\linewidth]{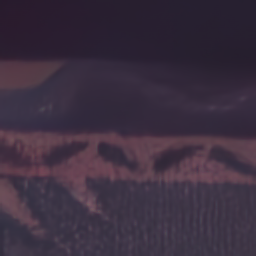}} &
		{\includegraphics[width=0.19\linewidth]{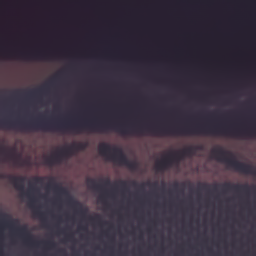}} &
		{\includegraphics[width=0.19\linewidth]{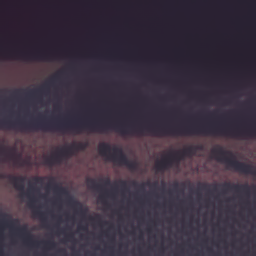}} & 
		{\includegraphics[width=0.19\linewidth]{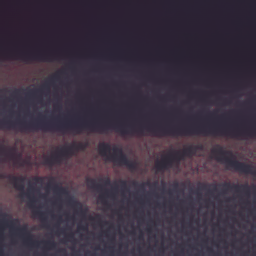}} & 
		{\includegraphics[width=0.19\linewidth]{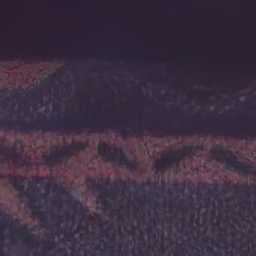}}  \\
		42.45/0.9560 & \textbf{42.56}/\textbf{0.9569} & 42.46/0.9557 & 41.98/0.9507 & 41.09/0.9407
	\end{tabular}
}
\caption{Examples of fine-tuning solutions in challenging scenarios. \textbf{(Best viewed with zoom-in)}}
\label{fig:finetune}
\end{figure}

\subsubsection{\textbf{Solutions Based on Fine-tuning}}
During the design phase, we deliberately avoid using black-box operations to improve performance. As a result, YOND exposes numerous intermediate results and adjustable parameters. Users can manually modify these explicit adjustable parameters to handle challenging cases.
To exhibit the fine-tuning solutions of YOND, we select a few representative adjustable parameters for instance in Figure~\ref{fig:finetune}.

The flat region mask in CNE serves as a representative adjustable parameter. As shown in Figure~\ref{fig:discuss_mask}, ATS may come out with similar scores at multiple quantiles when large flat regions are present, causing the optimal solution of Eq.~\eqref{eq:ATS} to deviate from the actual best choice. By adjusting the threshold based on ATS output, a better flat region mask can sometimes be obtained, as demonstrated by the difference between the ``Old Mask" and ``New Mask". For challenging scenarios, manual calibration of the flat region mask is also feasible.

Another representative adjustable parameter is the noise parameter estimated by CNE. Single-image noise estimation is typically inaccurate for images lacking flat regions. However, since the same camera follows the same noise model, noise parameters from images captured with identical camera settings can be adopted. For instance, the noisy image ``0183" shown in Figure~\ref{fig:failure} is captured by a Google Pixel camera at ISO-6400, and the SIDD training set includes a noisy image ``0108" captured under the same settings. Image ``0108" contains large flat regions, which is friendly to accurate noise estimation. As shown in Figure~\ref{fig:discuss_CNE}, applying the noise parameters estimated from image ``0183" to image ``0108" significantly enhances image detail. For datasets with stable noise parameters and sufficient data, we can calibrate the noise model of cameras using CNE, similar to the approach in Zou~\cite{CVPR22/FG-ELD}. According to our experiments, replacing the CNE-estimated noise parameters with calibrated ones yields a 0.1dB improvement on the LRID dataset.

Last but not least, the noise level $\sigma_\text{SNR}$ is the most representative adjustable parameter. In the main text, it is used to counteract the additional 3\% noise introduced by EM-VST. The noise level $\sigma_\text{SNR}$ directly controls the denoising strength of SNR-Net. Intuitively, increasing the noise level $\sigma_\text{SNR}$ further suppresses residual noise, while decreasing it releases more details. As shown in Figure~\ref{fig:discuss_snr}, reducing the noise level $\sigma_\text{SNR}$ results in denoised images with clearer details. However, it is essential to note that adjusting $\sigma_\text{SNR}$ only balances details and noise; it does not generate additional details.

\begin{figure}[t]
\footnotesize
\centering
\includegraphics[width=1.0\linewidth]{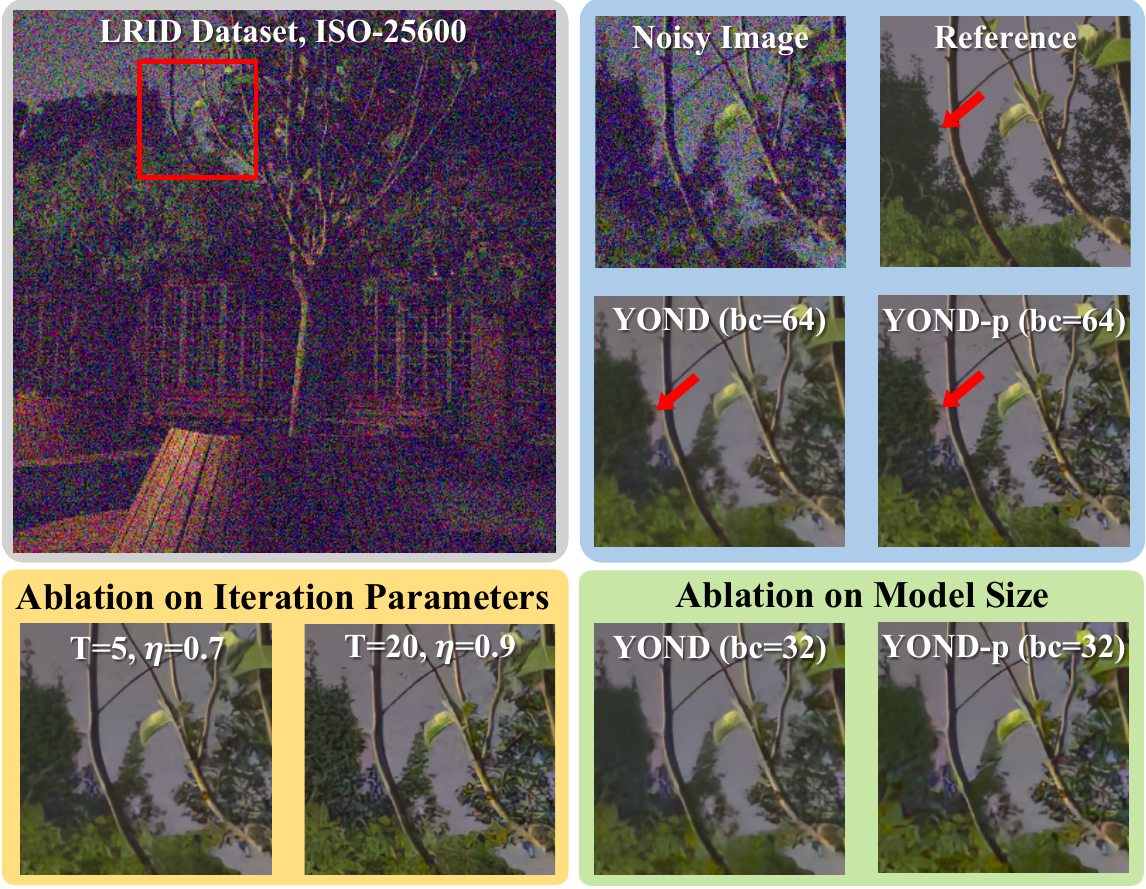}
\caption{A real-world use case of YOND-p, including ablations on iteration parameters and model size. \textbf{(Best viewed with zoom-in)}}
\label{fig:YOND-p}
\end{figure}

\subsubsection{\textbf{Solutions Based on YOND-p}}\label{sec:discuss_YOND-p}
Users consistently seek clearer denoising results, which may exceed the capacity of a single denoiser. YOND-p provides a solution by leveraging the schedule of generative models to achieve clearer results.

The blue region in Figure~\ref{fig:YOND-p} illustrates the performance of YOND-p in a challenging real-world scenario. The noisy image sourced from the LRID dataset is captured at ISO-25600 (ISO-6400 with 4$\times$ digital gain) and has a noise level of $\sigma_{\text{SNR}}$=32.87 after VST. Under the standard YOND strategy, even increasing the base channel count from 32 to 64 does not significantly improve clarity. In contrast, when we apply the iterative denoising strategy ($T$=10, $\eta$=0.8) proposed in Section~\ref{sec:YOND-p}, YOND-p generates rich textures in the leaf region. The textures indicated by the red arrows are evidently generated from the initial denoising result of YOND and differ notably from the reference image.

The yellow region in Figure~\ref{fig:YOND-p} presents our ablation studies on the iteration parameters. Here we adjust the iteration count $T$ and the noise decay factor $\eta$ to ensure the guiding value $\sigma_{t}$ is approximately 5 in the final denoising step. Halved iterations ($T$=5, $\eta$=0.7) yield fewer details but pleasing adaptive sharpening on edges. Doubled iterations ($T$=20, $\eta$=0.9) produce abundant details, however, these details appear somewhat artificial due to the model size being 20 times smaller than a classical diffusion model.

The green region in Figure~\ref{fig:YOND-p} shows the results of directly converting the YOND used in the main text to YOND-p. This version of YOND (bc=32) is identical to the one used in Figure~\ref{fig:teaser}. The YOND-p (bc=32) lacks sufficient generative capability, resulting in performance akin to simple sharpening. Under such a poor model size, an aggressive iteration strategy (\eg, $T$=20, $\eta$=0.9) not only fails to generate more details but also introduces numerous artifacts.

In conclusion, YOND-p enhances the flexibility of YOND, enabling the generation of rich details through our generative iteration strategy.

\subsection{Adapting to Noise Model Variations}\label{sec:noise_model}
In the main paper, we assume the camera noise follows a Poisson-Gaussian model, regarded as the most general model for describing the imaging noise in cameras~\cite{EMVA1288}. 
Moreover, handling the noise out of the Poisson-Gaussian model requires additional information, which generally conflicts with the premise of ``blind denoising". Therefore, the noise significantly out of the Poisson-Gaussian model is not the primary focus of this paper. Furthermore, the competitive denoising performance of YOND on low-light denoising datasets has demonstrated its robustness across minor variations in the noise model.

Although the noise out of the Poisson-Gaussian model is not our primary focus, we acknowledge the practicality of considering additional adaptation of the noise model, particularly in extreme low-light conditions or on specific sensors. Consequently, we are open to offering potential solutions for reference:

\begin{itemize}[leftmargin=0.7cm]
\item If paired real data is available, we can combine YOND with LED~\cite{ICCV23/LED} or PMN~\cite{TPAMI23/PMN}, fine-tuning SNR-Net to adapt the camera-specific data.

\item If the calibration materials are available, we can obtain the noise model of specific cameras via PNNP~\cite{PNNP}. 
After the noise model is known, we can compute the expectation term after GAT and modify the EM-VST to adapt the new noise model.

\item If the known noise model is significantly different from the Poisson-Gaussian model, we can synthesize the noise after GAT to train SNR-Net.

\item If the semantic content of noisy images is stable, we can train SNR-Net in a generative denoising manner~\cite{NIPS20/DDPM, ICLR21/DDIM, DMID, ICLR21/VE-SDE}, utilizing image priors to resist unknown noise.
\end{itemize}

We discover that the generative denoising approach has minimal conflict with ``blind denoising", which is the direction of our future work.

\section{Conclusion}
In this paper, we propose a novel method for blind raw image denoising free from camera-specific data dependency named YOND. Our method shows superior generalization to data from diverse unknown cameras once training on synthetic data.
YOND consists of three key modules: CNE, EM-VST, and SNR-Net.
CNE provides high precision in noise parameter estimation to ensure robust denoising performance. EM-VST exhibits low error in VST bias correction to ensure the exact color of denoised images. SNR-Net supports controllable raw AWGN denoising to deliver clear images under adaptive adjustment.
Extensive experiments on unknown cameras and flexible solutions for challenging cases demonstrate the superior practicality of our method.

\bibliographystyle{IEEEtran}
\bibliography{YOND}

\begin{IEEEbiography}[{\includegraphics[width=1in,height=1.25in,clip,keepaspectratio]{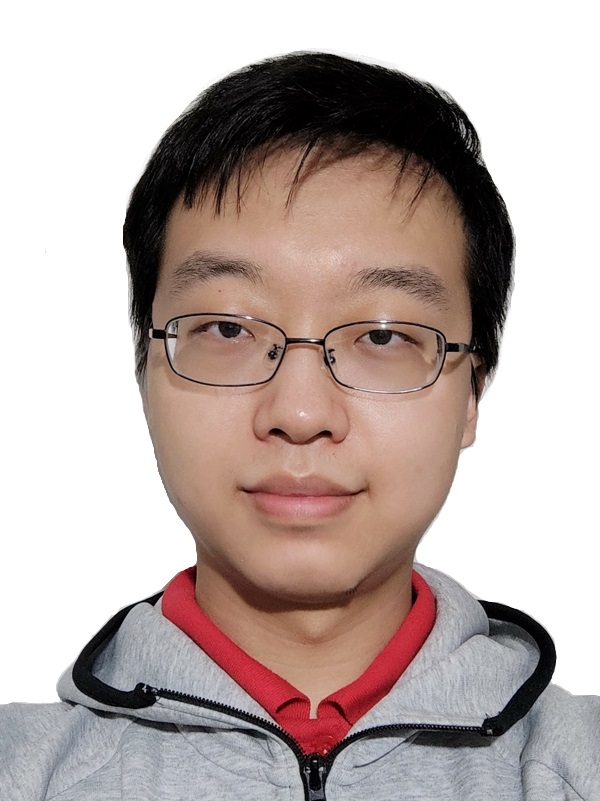}}]{Hansen Feng}
received the BS degree from the University of Science and Technology Beijing, China, in 2020. He is currently a Ph.D. student with the School of Computer Science and Technology, Beijing Institute of Technology. His research interests include computational photography and image processing. He received the Best Paper Runner-Up Award of ACM MM 2022.
\end{IEEEbiography}

\begin{IEEEbiography}[{\includegraphics[width=1in,height=1.25in,clip,keepaspectratio]{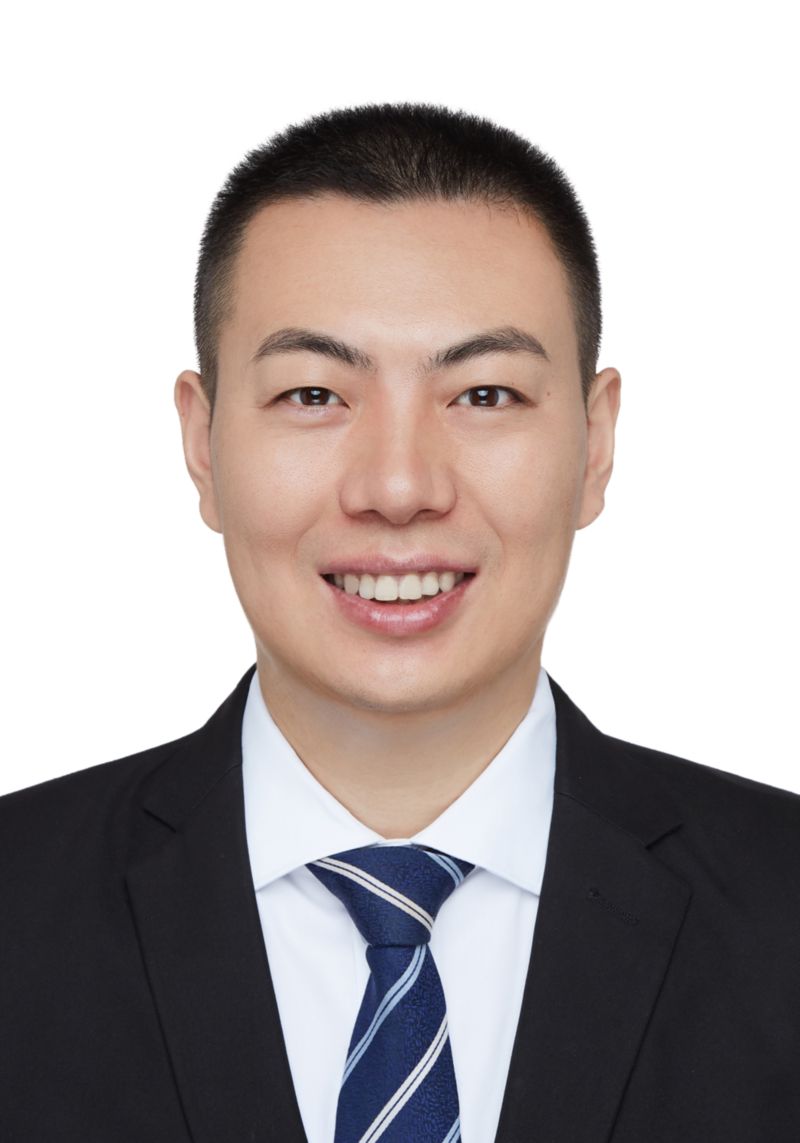}}]{Lizhi Wang}
(M’17) received the BS and PhD degrees from Xidian University, Xi’an, China, in 2011 and 2016, respectively. He is currently a professor in the School of Artificial Intelligence, Beijing Normal University. His research interests include computational photography and image processing. He is serving as an associate editor of IEEE Transactions on Image Processing. He received the Best Paper Runner-up Award of ACM MM 2022 and Best Paper Award of IEEE VCIP 2016.
\end{IEEEbiography}

\begin{IEEEbiography}[{\includegraphics[width=1in,height=1.25in,clip,keepaspectratio]{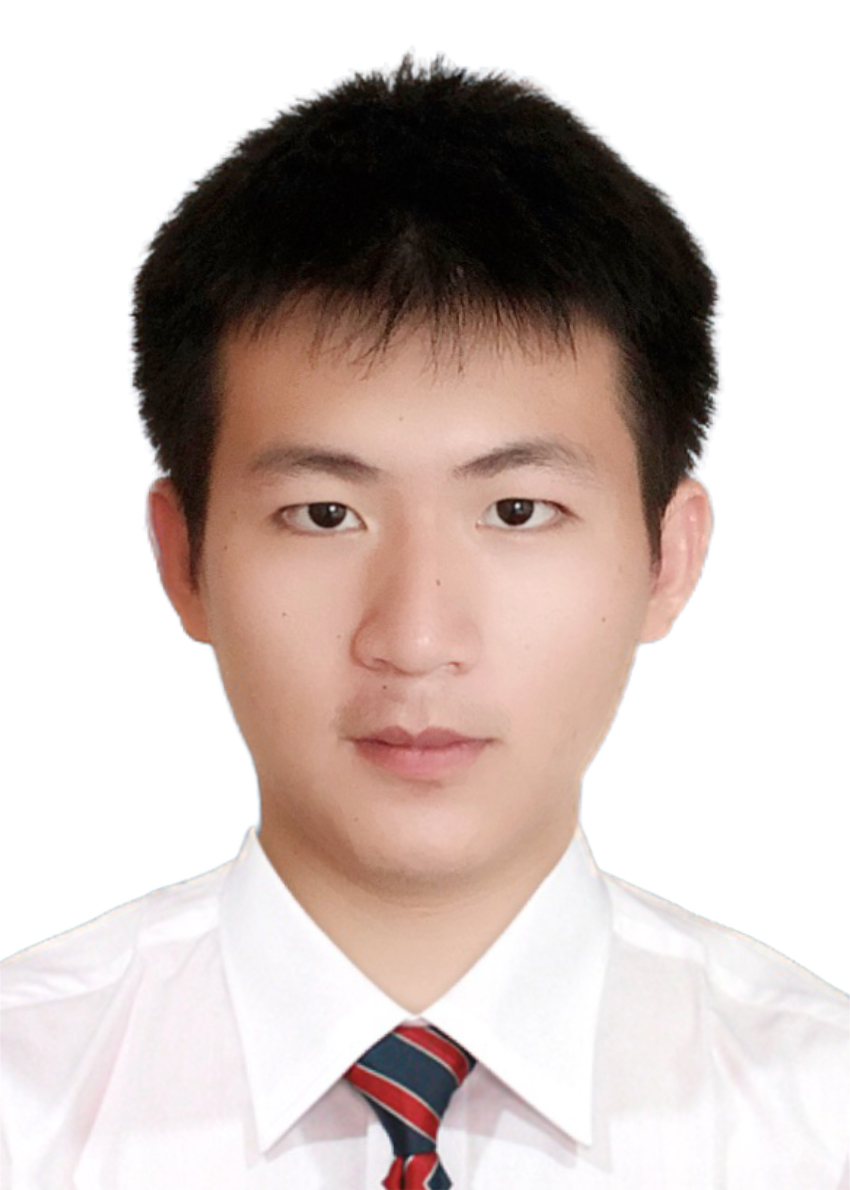}}]{Yiqi Huang}
received the B.S. degree from the school of Information Science and Engineering, Lanzhou University, Lanzhou, China, in 2022. He is currently a M.D. student with the School of Computer Science and Technology, Beijing Institute of Technology. His research interests include computational photography and image processing.
\end{IEEEbiography}


\begin{IEEEbiography}[{\includegraphics[width=1in,
	height=1.25in,
	clip,keepaspectratio]{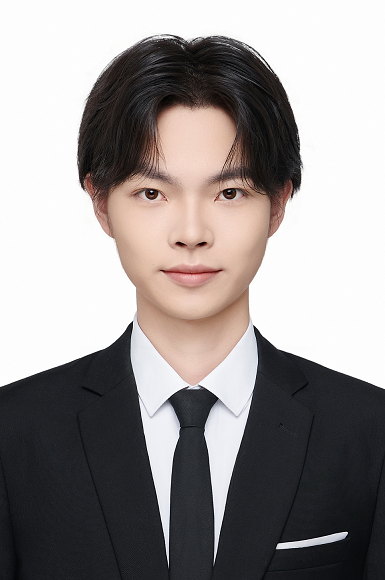}}]{Tong Li}
received the BS degree from Beijing Institute of Technology, China, in 2023. He is currently a Master student with the School of Computer Science and Technology at Beijing Institute of Technology. His research interests include computational photography and image processing.
\end{IEEEbiography}

\begin{IEEEbiography}[{\includegraphics[width=1in,height=1.25in,clip,keepaspectratio]{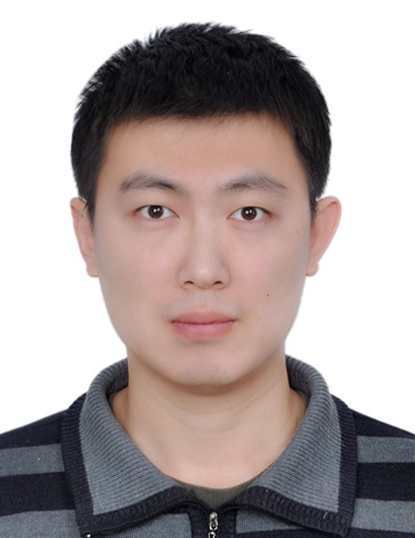}}]{Lin Zhu}
(M'22) received the B.S. degree in computer science from Northwestern Polytechnical University, Xi’an, China, in 2014, the M.S. degree in computer science from North Automatic Control Technology Institute, Taiyuan, China, in 2018, and the Ph.D. degree from the School of Electronics Engineering and Computer Science, Peking University, Beijing, China, in 2022.,He is currently an Assistant Professor with the School of Computer Science, Beijing Institute of Technology, Beijing. His current research interests include image processing, neuromorphic computing, and spiking neural network.
\end{IEEEbiography}

\begin{IEEEbiography}[{\includegraphics[width=1in,height=1.25in,clip,keepaspectratio]{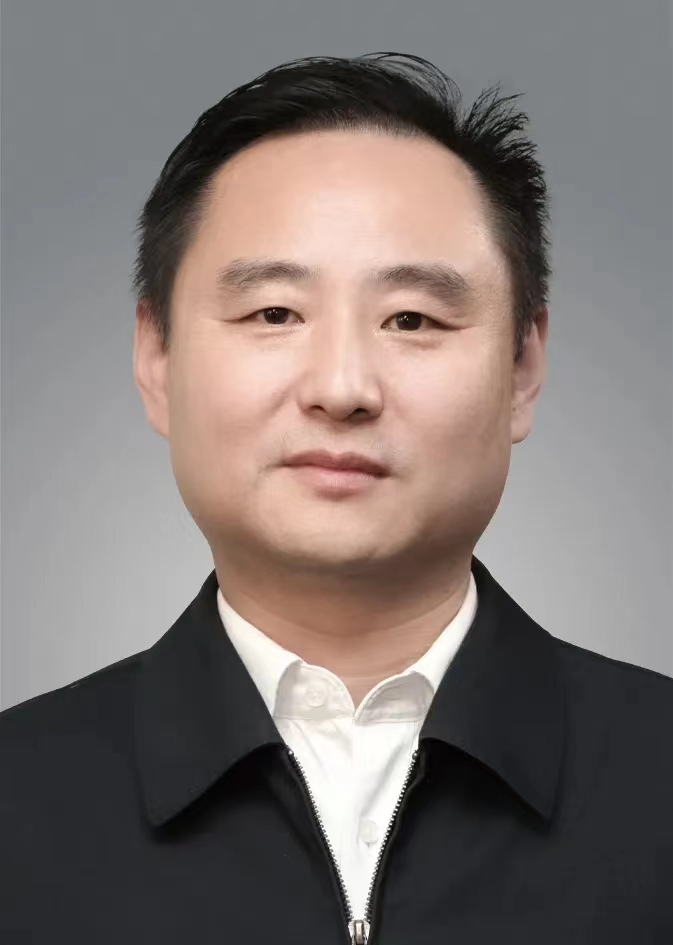}}]{Hua Huang}
(SM’19) received the BS and PhD degrees from Xi’an Jiaotong University, in 1996 and 2006, respectively. He is currently a professor in the School of Artificial Intelligence, Beijing Normal University. He is also an adjunct professor with Xi’an Jiaotong University and Beijing Institute of Technology. His main research interests include image and video processing, computational photography, and computer graphics. He received the Best Paper Award of ICML2020/EURASIP2020/ PRCV2019/ChinaMM2017.
\end{IEEEbiography}

\vfill

\end{document}